\documentclass[letterpaper]{article} % DO NOT CHANGE THIS
\usepackage[submission]{aaai25}  % DO NOT CHANGE THIS
\usepackage{times}  % DO NOT CHANGE THIS
\usepackage{helvet}  % DO NOT CHANGE THIS
\usepackage{courier}  % DO NOT CHANGE THIS
\usepackage[hyphens]{url}  % DO NOT CHANGE THIS
\usepackage{graphicx} % DO NOT CHANGE THIS
\usepackage{natbib}  % DO NOT CHANGE THIS AND DO NOT ADD ANY OPTIONS TO IT
\usepackage{caption} % DO NOT CHANGE THIS AND DO NOT ADD ANY OPTIONS TO IT
\usepackage{algorithm}
\usepackage{algorithmic}
\usepackage{amsmath}
\usepackage{multirow}
\usepackage{pifont}
\usepackage{amsfonts}
\usepackage[flushleft]{threeparttable}
\usepackage{newfloat}
\usepackage{listings}
\DeclareCaptionStyle{ruled}{labelfont=normalfont,labelsep=colon,strut=off} % DO NOT CHANGE THIS
\floatstyle{ruled}
\newfloat{listing}{tb}{lst}{}
\floatname{listing}{Listing}
\title{STEAM: Squeeze and Transform Enhanced Attention Module}

\author {
    Rishabh Sabharwal\textsuperscript{\rm 1},
    Ram Samarth B B\textsuperscript{\rm 1},
    Parikshit Singh Rathore\textsuperscript{\rm 1},
    Punit Rathore\textsuperscript{\rm 1}
}
\affiliations {
    \textsuperscript{\rm 1}Indian Institute of Science, Bangalore, Karnataka, India\\
    \{srishabh, ramsamarthbb, parikshits, prathore\}@iisc.ac.in
}

\usepackage{bibentry}
\begin{document}

\maketitle
% Welcome 
\begin{abstract}
Channel and spatial attention mechanisms introduced in earlier work enhance the representational capabilities of deep convolutional neural networks (CNNs) but often increase parameter and computational costs. While recent approaches focus solely on efficient feature context modeling for channel attention, we aim to model both channel and spatial attention comprehensively with minimal parameters and reduced computation. Leveraging the principles of relational modeling in graphs, we introduce a constant-parameter module, \textit{STEAM: Squeeze and Transform Enhanced Attention Module}, which integrates channel and spatial attention to enhance the representation power of CNNs. To our knowledge, we are the first to propose a graph-based approach for modeling both channel and spatial attention, utilizing concepts from multi-head graph transformers. Additionally, we introduce \textit{Output Guided Pooling} (OGP), which efficiently captures spatial context to further enhance spatial attention. We extensively evaluate STEAM for large-scale image classification, object detection and instance segmentation on standard benchmark datasets. STEAM achieves a \(2\%\) increase in accuracy over the standard ResNet-50 model with only a meager increase in GFLOPs. Furthermore, STEAM outperforms the leading modules, ECA and GCT, in terms of accuracy while achieving a threefold reduction in GFLOPs. The code will be made available upon acceptance.

%STEAM is a constant-parameter module that can be seamlessly integrated with deep CNN architectures.

%Therefore, we belive that STEAM establishes a new benchmark for improving channel and spatial attention, positioning it as the state-of-the-art.

%For instance, in ResNet-50, ECA and GCT add an additional 1.1e-2 GFLOPs each, while STEAM adds a minimal 0.357e-2 GFLOPs.

% We present STEAM: Squeeze and Transform Enhanced Attention Module, inspired from graph representation learning. By leveraging the fundamental principles of relational modeling in graphs, we introduce effective channel and spatial graph attention modules. Our modules utilize multi-head attention mechanism to model both channel and spatial attention, offering a parameter and computation-efficient solution with enhanced feature context and relational modeling. To the best of our knowledge, we are the first to propose a constant-parameter module independent of the number of channels. We achieve a \(2\%\) increase in accuracy over the standard ResNet-50 model with only an additional 3.57e-3 GFLOPs. Our STEAM unit is highly efficient in terms of GFLOPs, when compared to ECA and GCT, which are currently among the simplest effective modules. We extensively evaluate STEAM for large scale image classification, object detection and instance segmentation. By presenting STEAM, we introduce a new SOTA attention module informed by graph representation learning.
\end{abstract}

% Uncomment the following to link to your code, datasets, an extended version or similar.
%
% \begin{links}
%     \link{Code}{https://aaai.org/example/code}
%     \link{Datasets}{https://aaai.org/example/datasets}
%     \link{Extended version}{https://aaai.org/example/extended-version}
% \end{links}

\section{Introduction}
% CBAM paper \cite{woo2018cbam} 
% Attention mechanisms have revolutionized modern neural networks, evolving from human visual perception principles, neural machine translation \cite{bahdanau2014neural} to becoming integral in advanced models like Transformers\cite{vaswani2017attention}, enhancing performance by enabling precise focus on critical elements.
\textit{Convolutional Neural Networks} (CNNs) have greatly advanced vision tasks due to their strong representation capabilities. However, since convolutions operate within local spatial neighborhoods, repeated local operations are required to capture large receptive fields, leading to computational and optimization challenges \cite{wang2018non}.
% Convolutional Neural Networks (CNNs) excel in vision tasks but struggle in capturing long-range dependencies \cite{wang2018non}. Emphasizing repeated local operations to achieve large receptive fields introduces computational and optimization issues.
This limitation has driven the exploration of channel and spatial attention mechanisms to capture these dependencies better.
 Spatial Transformer Networks \cite{jaderberg2015spatial} use an adaptive mechanism to enhance spatial dependencies and NLNets \cite{wang2018non} utilize non-local operators to capture long-range dependencies. SENet \cite{hu2018squeeze} uses adaptive recalibration of feature maps using bottleneck \textit{multi-layer perceptrons} (MLPs) to capture channel attention. 
 % NLNets \cite{wang2018non} utilized non-local operators to capture long-range dependencies  effecti spatial attention.
 BAM~\cite{park2018bam} and CBAM \cite{woo2018cbam} focus on integrating channel and spatial attention to improve dependency capture.
 % \cite{hu2018gather, ruan2021gaussian} focus on capturing effective feature context, while \cite{lee2019srm, jiang2024mca} emphasize higher-order moments to enhance channel attention. \cite{wang2018non, chen20182} have worked on capturing effective spatial attention. BAM \cite{park2018bam}, CBAM \cite{woo2018cbam} and GCNet \cite{cao2019gcnet} integrate both channel and spatial attention mechanisms to enhance the performance further.

\begin{table}[htbp]
	\small
	\centering
            \begin{threeparttable}
		\begin{tabular}{c|c|c|c|c}
			\hline
			Modules & DR & CA & SA  & Parameters 
		 \\ \hline 
			SE & \ding{51} & \ding{51}  & \ding{55} & $\frac{2}{r} \sum_{s=1}^{S}N_{s}\cdot C_{s}^{2}  $   \\ 
			% SRM & \ding{55} & \ding{51} & \ding{55} & $\sum_{s=1}^{S}N_{s}\cdot C_{s} *6 $ \\ 
                GCT & \ding{55} & \ding{51} & \ding{55} & $\sum_{s=1}^{S}N_{s}$ \\
			ECA & \ding{55} & \ding{51} & \ding{55} & $ \sum_{s=1}^{S}N_{s}\cdot |( log_{2}C +1) /2 |_{odd}$ \\ 
			CBAM & \ding{51} & \ding{51} & \ding{51} & $\sum_{s=1}^{S}N_{s}\cdot (C_{s}^{2} *\frac{2}{r} + {\text{k}^2})  $ \\ 
			% SA & \ding{55} & \ding{51} & \ding{51} & $\sum_{s=1}^{S}N_{s}\cdot C_{s} *\frac{3}{G}$\\ 
                % MCA & \ding{55} & \ding{51} & \ding{55} & $\sum_{s=1}^{S}N_{s}\cdot C_{s} * 2$ \\
                STEAM & \ding{55} &  \ding{51} & \ding{51} & Independent of C!\\ \hline
			
		\end{tabular}
        % \caption{Number of Parameters for different modules. DR, CA, SA denotes dimensionality reduction, channel attention, spatial attention respectively. Here \ding{51}, \ding{55} means presence or absence}
        \caption{Comparison with SOTA modules based on channel dimensionality reduction (DR), Channel Attention (CA) and Spatial Attention (SA), total number of parameters.}
        % \(C\) denotes the number of channels, \(r\) represents the reduction ratio of SE and CBAM, \(|\cdot|\) indicates the nearest odd number, \(N_{s}\) denotes the number of repeated blocks in stage \(s\), and \(S\) represents the total number of stages. \(\text{k}\) denotes the size of 2-D kernel used in CBAM.}
        \label{table-param}
        \begin{tablenotes}
        \item \(C\) -- the number of channels; \(r\) -- the reduction ratio of SE and CBAM; \(N_{s}\) -- the number of repeated blocks in stage \(s\); \(S\) -- the total number of stages; and \(\text{k}\) -- the size of 2-D kernel used in CBAM. \(|\cdot|\) indicates the nearest odd number.
        \end{tablenotes}
        \end{threeparttable}
\end{table}

The SE block in SENet captures inter-channel interactions for efficient attention but significantly increases the number of parameters. Moreover, the dimensionality reduction in fully connected layers fails to capture direct dependencies, as highlighted by ECA \cite{wang2020eca}. This issue also affects methods such as CBAM \cite{woo2018cbam} and GE \cite{hu2018gather}, which employ SE-like dimensionality reduction techniques for modeling channel attention. 
% Additionally, CBAM’s pixel-by-pixel spatial attention imposes an unnecessary computational burden on low-level features in the model’s early stages. 
Methods such as GCT \cite{ruan2021gaussian} utilize non-parametric strategies, whereas SRM \cite{lee2019srm} adopts moment-based techniques to enrich feature context and boost channel attention. 

% SGE \cite{li2019spatial} and SANet \cite{zhang2021sa} improve channel and spatial attention by employing group convolution strategies.
% Subsequently, ECA \cite{wang2020eca} struggles to capture cross-channel attention effectively as the receptive field remains limited even with an increased number of channels.
% In grouped convolution techniques, SGE \cite{li2019spatial} employs spatial mechanisms on grouped channel sub-features but inadequately exploits spatial-channel correlations, limiting efficiency. SANet \cite{zhang2021sa} introduces Shuffle Attention, but it’s unclear if its efficiency in capturing channel and spatial attention stems from the proposed method or merely the grouped operations.
\begin{figure*}
    \centering
    \includegraphics[width=\textwidth]{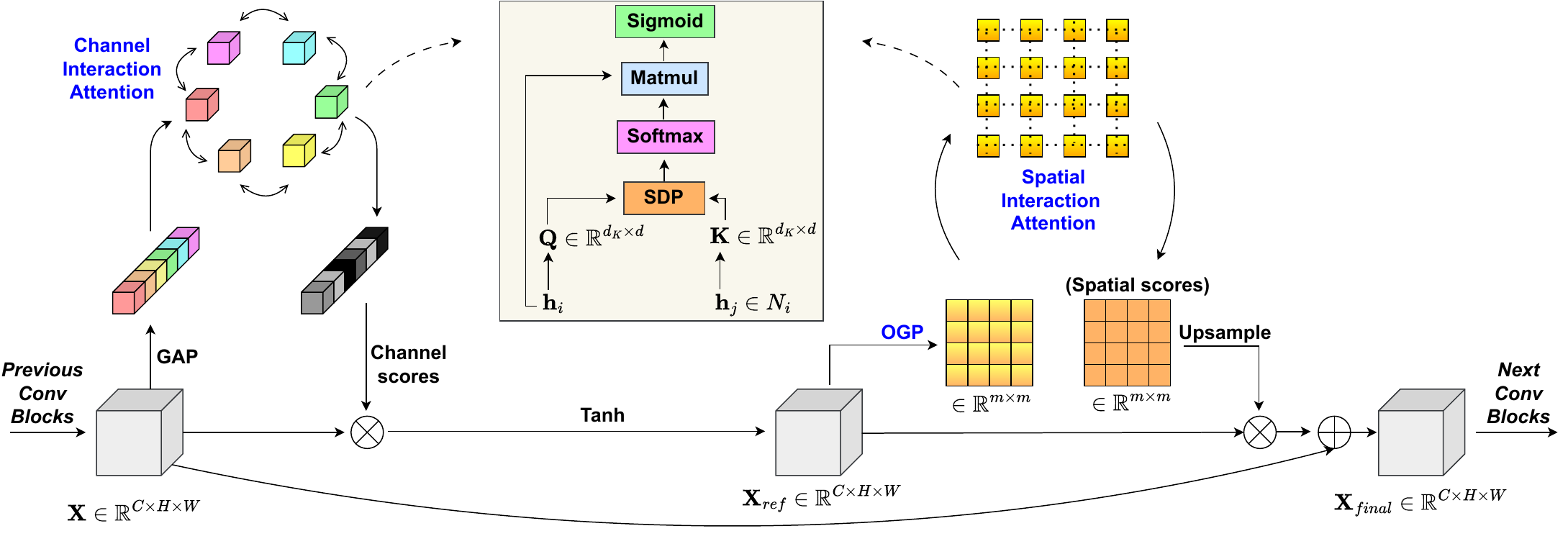}
    \caption{Detailed Overview of STEAM.} 
    \label{fig:mainarchitecture}
\end{figure*}

% Assessing these studies, we observe a focus on implementing attention mechanisms, although this often results in a higher number of parameters. Various methods enhance features, some employ efficient backbone-inspired techniques, and others aim to reduce parameters while targeting specific types of attention.
% Upon assessing these studies, we observe that most approaches substantially increase the number of parameters to improve performance, 
% % Some techniques leverage efficient backbone-inspired methods to reduce computations, 
% while others aim to reduce parameters by focusing only on channel attention as evident in Tabel \ref
Upon assessing these studies, we find that many approaches significantly increase the number of parameters to enhance performance, while others focus on reducing parameters by concentrating solely on channel attention, as shown in Table \ref{table-param}.
% Analyzing the above shortcomings, we aim to understand the root for an effective solution. Current methods enhance cross-channel attention to help CNNs determine ‘what’ to attend to by capturing feature context or learning complex interactions. These approaches often increase parameters and computational burden, heavily relying on MLP mechanisms. Efforts to reduce complexity or improve parameter efficiency frequently compromise effectiveness. Similarly, spatial attention methods face analogous issues.
% We build upon the principles of graph relation modeling and recent advances in graph representation learning to develop our method. From channel perspective, each channel, represented by its spatial map can be viewed as a node associated with its feature vector, pushing us towards channel graph. From a spatial perspective, each spatial unit, which encapsulates positional and interaction details, can be viewed as a node with the channel’s global information as its feature, guiding us towards spatial modeling. 
In contrast, our method employs graph relational modeling to effectively capture both channel and spatial attention, striking a balance between performance and computational efficiency.
% Our approach employs graph relational modeling to effectively capture attention, while striving to minimize computational and parameter burdens. 
% From a channel perspective, each channel, represented by its corresponding spatial map, can be viewed as a node with the global information of this spatial map as its feature vector. 
From the channel perspective, each channel can be viewed as a node with the global information of its corresponding spatial map as the feature vector.
From the spatial perspective, each spatial unit, encompassing positional and interaction details, can be perceived as a node with global information across all the channels as its attribute. Therefore, channel and spatial attention can be modeled as graph attention mechanisms, which we refer to as \textit{Channel Interaction Attention} (CIA) and \textit{Spatial Interaction Attention} (SIA).
% CIA and SIA utilize multi-head scaled dot-product attention to improve model representation capabilities. 
We introduce a new \textit{Output Guided Pooling} (OGP) to capture spatial context efficiently which enhances our SIA. We name our approach that integrates both CIA and SIA as STEAM: \textit{Squeeze and Transform Enhanced Attention Module}.

% Table \ref{table-param} summarizes existing attention modules based on (i) whether they incorporate dimensionality reduction (DR), (ii) their ability to capture Channel Attention (CA) and Spatial Attention (SA), and (iii) their total parameter count. As observed, STEAM effectively leverages both channel and spatial attention without introducing dimensionality reduction. To evaluate our method, we perform large-scale image classification on ImageNet, as well as object detection and instance segmentation on MS COCO. STEAM integrated with ResNet-50 adds a minimal 320 parameters and 3.57e-3 GFLOPs, yielding an increase of \(2\%\) Top-1 accuracy over standard ResNet-50. 
As evident in Table \ref{table-param}, STEAM leverages both channel and spatial attention without introducing dimensionality reduction in a parameter-efficient manner. Furthermore, STEAM integrates seamlessly with various network architectures, including ResNets \cite{he2016deep, xie2017aggregated} and lightweight models such as ShuffleNet \cite{zhang2018shufflenet} and ShuffleNet-V2 \cite{ma2018shufflenet}.

In general, the key contributions of this work can be summarized as follows: 
% (1) We conceptualize channel and spatial attention utilizing a graphical approach by introducing channel and spatial graphs, aiming for efficient representation learning. 
(1) We develop channel and spatial attention through a graphical approach by defining channel and spatial graphs, with the goal of achieving efficient representation learning.
(2) We use multi-head attention, inspired by graph transformers, to capture various relationships within channel and spatial graphs. Additionally, we introduce \textit{Output Guided Pooling} for optimal spatial attention modeling. (3) We develop a constant-parameter module independent of the backbone it is integrated with. (4) We perform large-scale image classification on ImageNet, as well as object detection and instance segmentation on MS COCO dataset. Our experimental results demonstrate that STEAM outperforms current \textit{state-of-the-art} (SOTA) modules, while adding minimal parameters and GFLOPs, highlighting its efficiency in terms of parameters and computation. For instance, STEAM integrated with ResNet-50 adds a minimal 320 parameters and 3.57e-3 GFLOPs, yielding an increase of \(2\%\) Top-1 accuracy over standard ResNet-50.

\section{Related Works}
\textbf{Attention Mechanism.} Attention mechanisms, enhancing deeper CNNs, have been broadly explored under channel, spatial, and combined channel-spatial attention. \cite{zagoruyko2016paying} introduced spatial attention function for encoding the input areas focused by the network for decision-making. SENet \cite{hu2018squeeze} modeled channel attention by recalibrating feature maps through squeeze and excitation operations. \cite{roy2018recalibrating} achieves consistent segmentation gains by incorporating three variants of SE modules with CNNs. GE \cite{hu2018gather} uses a gather-excite operator pair to aggregate contextual information and modulate feature maps. GCT \cite{ruan2021gaussian} uses a Gaussian function to excite global contexts, removing the need for parameterized contextual feature transforms. GSoP \cite{gao2019global} employs global second-order pooling for holistic image representation, SRM \cite{lee2019srm} utilizes style-based pooling. Residual Attention Networks \cite{wang2017residual} enhance feature maps and increase noise resilience by computing robust 3D attention maps. CBAM \cite{woo2018cbam} builds on this by independently modeling and sequentially combining channel and spatial attention mechanisms. ECA-Net \cite{wang2020eca} uses 1-D convolution to capture local cross-channel interactions, addressing dimensionality reduction issues. Approaches such as SGE \cite{li2019selective} and SANet \cite{zhang2021sa} leverage multi-branch architecture designs similar to ShuffleNet \cite{zhang2018shufflenet} and ShuffleNet-V2 \cite{ma2018shufflenet} to integrate attention mechanisms.
% SGE \cite{li2019selective} enhances feature maps by dividing channels into groups and applying lightweight spatial attention. 
% SANet \cite{zhang2021sa} combines spatial and channel attention using Shuffle Units and a channel shuffle operator to effectively aggregate features.
NLNet \cite{wang2018non} employs non-local operations to capture long-range dependencies through global interactions, similar to self-attention. GCNet \cite{cao2019gcnet} unifies non-local and SE blocks into a \textit{Global Context} (GC) block for enhanced context modeling. A\(^\textit{2}\)-Net \cite{chen20182}  captures long-range feature interdependencies by employing universal gathering and distribution mechanisms.
\\
\textbf{Graph Neural Networks (GNNs)}. GNNs \cite{scarselli2008graph, he2016deep} model complex graph data using message passing \cite{gilmer2017neural}. Advances include GCN \cite{kipf2016semi}, GAT \cite{velivckovic2017graph}, and GraphSAGE \cite{hamilton2017inductive}, each employing unique aggregation techniques. The relational inductive bias of GNNs, as explained by \cite{battaglia2018relational}, makes them superior to MLPs for graph data. Attention mechanisms in GNNs have garnered significant interest: \cite{brody2021attentive} improved GAT with dynamic attention,  \cite{dwivedi2020generalization} introduced Graph Transformers with positional encodings, and models like SAN \cite{kreuzer2021rethinking}, and GPS \cite{rampavsek2022recipe} enhanced positional and structural encoding. The application of GNNs in vision is well-reviewed in \cite{jiao2022graph}.
Vision GNN \cite{han2022vision}, utilizes an approach similar to ViT\cite{dosovitskiy2020image}, in which each image patch is considered as a graph node that is linked to its nearest neighbors.
\cite{aflalo2023deepcut} utilized graph spectral clustering for unsupervised image segmentation.
% , leveraging intermediate features from pre-trained ViT \cite{dosovitskiy2020image} to build a correlation graph.
% \cite{xu2024gtp} introduced Graph-based Token Propagation to balance efficiency and information preservation in ViT\cite{dosovitskiy2020image}.

%% Beginning with the main methodology
\section{Proposed Method}
% This section details our preliminary na\"ive methods for graph-based channel attention, explains the development of both our Channel Interaction Attention (CIA) and Spatial Interaction Attention (SIA) modules, and subsequently integrates these to create STEAM. The schematic diagram of STEAM is shown in  Figure~\ref{fig:mainarchitecture}.
This section provides an in-depth explanation of our Channel and Spatial Interaction Attention (CIA and SIA) mechanisms. We also describe the construction of our proposed method STEAM using these mechanisms, with a schematic diagram shown in Fig.~\ref{fig:mainarchitecture}.

% \subsection{Towards an Initial Channel Interaction Mechanism}
\subsection{Our initial steps towards a na\"ive Channel Attention}

Building on ECA’s insights, we forego the dimensionality reduction strategies used in SENet, CBAM, and GE, opting for direct cross-channel attention modeling. Given the intermediate feature map \(\textbf{X} \in \mathbb{R}^{C \times H \times W}\), we treat each channel as a node, with its spatial map \(\textbf{x}_{c} \in \mathbb{R}^{H*W}\) as its feature vector. We derive a correlation graph using channel-wise correlation matrix \(\textbf{A} = \textbf{XX}^{T} \in \mathbb{R}^{C \times C}\) computed from node features. But as CNNs deepen, the number of channels increases exponentially, resulting in large and dense graphs that pose computational challenges. Modeling such graphs requires multiple GNN layers to increase the receptive field.  However, deep GNNs tend to oversmooth \cite{nt2019revisiting,oono2019graph}, making node representations very similar, thereby degrading the performance. To avoid such dense graphs, we experimented with \(k\)-NN graphs by selecting the top \(k\) neighbors based on similarity scores, a strategy used in methods like LLE \cite{roweis2000nonlinear} to infer graphs. However, since we aim to model channel importance rather than similarity, \(k\)-NN graphs
were found to be ineffective\footnote{refer to Table 10 in Appendix.}. Additionally, different correlation matrices for each image lead to excessive memory usage, hindering graph representation learning.

\subsection{Exploring our Channel Interaction Mechanism}
% We realized the importance of comprehending channel graph construction at a fundamental level after facing initial shortcomings.
In ECA's 1-D cross-channel convolution, the kernel size defines the local receptive field, where weighted aggregation corresponds to convolution. Drawing this analogy to graphs, the kernel size corresponds to a node’s neighborhood, with convolution modeled as message passing within this neighborhood. Inspired by these correlations, we design a channel graph \(G_{\text{c}}\) in which each channel is linked to its immediate 1-hop neighbors. Furthermore, the first channel is connected to the last one, creating a cyclic graph (Fig.~\ref{fig:mainarchitecture}).
% To enhance graph representation learning for channels, we initially work on constructing a Channel Graph.  the graph structure by connecting each channel to its immediate neighbors, with the first and last channels also linked resulting in a cyclic graph (see Fig. X). This approach finds its inspiration from 1-D convolutional cross channel attention as in ECA, where a kernel aggregates information within its region. This setup represents a node interacting with its local neighborhood, with an additional connection capturing long-range interactions.
Therefore, \(G_{\text{c}} = (V_{c}, E_{c})\) denotes our channel graph with \(C\) nodes and edges, where \(V_{c}\) and \(E_{c}\) represent the vertex and edge sets respectively. \(\textbf{X}_{c} \in \mathbb{R}^{C \times 1}\) represents the initial node (channel) features obtained by applying \textit{global average pooling} (GAP), a mapping \(\mathbb{R}^{C \times H \times W} \mapsto \mathbb{R}^{C \times 1 \times 1}\). Given the channel graph and its associated node features, channel attention can now be effectively modeled using graph attention mechanisms. Using scaled dot-product attention \cite{vaswani2017attention} over traditional self-attention \cite{bahdanau2014neural}, as implemented in GAT \cite{velivckovic2017graph}, has shown to be more effective empirically. We name this entire mechanism as \textit{Channel Interaction Attention }(CIA) which can be computed as  \cite{vaswani2017attention}:
% % We define “Channel Attentional Flow” (CAF) as an attention scoring mechanism, where updated node representations represent each channel’s attentional scores.  This results in a channel interaction graph, \(G_{channel}\) with \(|C|\) nodes and edges. To establish effective initial node features for each channel, we apply \textit{ global average pooling} (GAP) , a mapping \(\mathbb{R}^{H \times W} \mapsto \mathbb{R}^{1 \times 1}\), \(\forall c \in C\)  to aggregate spatial information. 
% Leveraging scaled dot product attention \cite{vaswani2017attention} over traditional self-attention \cite{bahdanau2014neural} mechanism used in GAT \cite{velivckovic2017graph} have proven to been empirically effective. CAF can be computed as

\begin{equation}
\textbf{A}_{i,j}^{h} = \text{softmax}\left(\frac{(\textbf{W}_{K}\textbf{X}_{c_{i}})^{T}\textbf{W}_{Q}\textbf{X}_{c_{j}}}{d_{K}}\right), \, \forall j \in N_{i} 
\label{eq1}
\end{equation}
\begin{equation}
\textbf{A} = \frac{1}{H}\sum_{h=1}^{H} \textbf{A}^{h} \, \, ,  \, \, \textbf{X}_{c_{att}} = \textbf{A}\textbf{X}_{c}
\label{eq2}
\end{equation}
\begin{equation}
   \alpha_{c} = \sigma(\textbf{X}_{c_{att}}),  \, \, \, \textbf{X}_{ref} = \alpha_{c} \otimes \textbf{X}
\label{eq3}
\end{equation}

Here, \(\textbf{W}_{K} , \textbf{W}_{Q} \in \mathbb{R}^{d_{K} \times d_{in}}\), (\(d_{in} = 1\)), are learnable projection matrices, where \(d_{K}\) specifies the dimensionality of the key and query vectors , \(d\) represents model's hidden dimension. \(\textbf{A}_{i,j}^{h}\) represents the attention score between node \textit{i} and its neighbor \(j\) in the \(h\)-th\footnote{Here \(h\) represents one of the attention heads. It is not to be confused with the node representation vector denoted by \(\textbf{h}_{i}\) in Fig.~\ref{fig:mainarchitecture}}. head among total \(H\) heads and \(N_{i}\) denotes the neighborhood of node \(i\). The final attention matrix \(\textbf{A}\) is computed by averaging the attention scores across all \(H\) heads. The updated representation for each channel is denoted by \(\textbf{X}_{c_{att}}\) which is further processed through a sigmoid activation (\(\sigma\)) to produce channel scores, \(\alpha_{c} \in \mathbb{R}^{C \times 1}\). During element-wise multiplication (denoted by \(\otimes\) in Fig.~\ref{fig:mainarchitecture} and Eq.~\ref{eq3}), \(\alpha_{c}\) is broadcasted across the spatial dimension to get the modified intermediate feature map \(\textbf{X}_{ref}\).
% Our CAF module offers key advantages: (i) Unlike ECA’s 1D convolution, CAF performs attentional convolution. (ii) CAF introduces multihead attention, enhancing the capture of complex interactions without computational overhead. While ECA uses adaptive kernel sizes, our relational attention with a neighborhood size of 2 enhances localized attention. Additionally, its similarity to information flow diagrams ensures effective global attention capture.
% CIA offers the following key advantages: (I) It uses graph attentional message passing over ECA’s 1-D convolution within the local receptive field (kernel/neighborhood), improving the modeling of relative channel importance. (II) It incorporates multihead attention to capture diverse relationship aspects, thereby stabilizing the learning process.
% (III) The relational inductive bias of GNNs over standard MLPs ensures robust generalization to unseen data \cite{battaglia2018relational}.

\subsection{Need for Spatial Attention}
% Begin Spatial interactional 
% -> Keys points
% - Why Spatial -> cite CBAM ? ICLR 2017 (spatial attention maps along channel)...
% - Output guided pooling -> 2 papers given by Rishabh 
% - CBAM computation of each cell
% - Why realtional spatial
% - ICLR 2017 for the formulation
% - why fixed size -> higher order -> pixel wise -> higher layers -> fix it to k x k..
% - equations cite
% - novel edge drop -> taking care of oversmoothing
% - try to cpature the effective spatial structure  
% - advantage
% - combine them squentaill -> justification(We have)
% Spatial attention is essential for identifying relevant features by highlighting spatial context \cite{woo2018cbam}. 

\cite{zagoruyko2016paying} emphasizes that spatial attention is crucial for encoding key spatial regions that influence a network’s output decision.
They propose a \textit{spatial attention mapping function} (SAMF) formulated as \(\mathbb{R}^{C \times H \times W} \mapsto \mathbb{R}^{H \times W}\) utilizing channel statistics. Early-layer neurons in the network activate strongly for low-level gradients, intermediate layers for distinctive features, and top layers for entire objects. Additionally, \cite{islam2020much} specifies that CNNs naturally extract positional information, with early layers capturing fine details and deeper layers focusing more on category-specific features.
% \cite{zagoruyko2016paying} states that spatial attention is essential for effectively encoding the key spatial regions that influence the network’s output decision. They define a \textit{spatial attention mapping function} (SAMF) as \(\mathbb{R}^{C \times H \times W} \mapsto \mathbb{R}^{H \times W}\) leveraging channel statistics to adaptively highlight specific image areas. Specifically, early-layer neurons show strong activation for low-level gradients, intermediate layers for distinctive features, and top layers for the entire object.
% Moreover, \cite{islam2020much} specifies that CNNs naturally extract positional information: early layers capture fine details, while deeper layers generate more abstract, category-specific features.  
% Moreover, convolutional layers maintain precise spatial location details \cite{kayhan2020translation}.
% \cite{zagoruyko2016paying} defines a \textit{spatial attention mapping function} (SAMF) as \(\mathbb{R}^{C \times H \times W} \mapsto \mathbb{R}^{H \times W}\) leveraging channel statistics. 
% These maps adaptively highlight specific image areas. Early-layer neurons show strong activation for low-level gradients, intermediate layers for distinctive features, and top layers for the entire object. 
This inherent positional and spatial context in CNNs motivates us to use GNNs for modeling spatial attention, as they effectively capture both positional and relational information
% This inherent positional and spatial context in CNNs motivates us to employ GNNs for spatial attention modeling, as GNNs effectively capture both positional and relational information. 
Therefore, we construct an initial spatial graph from the spatial attention map obtained by using SAMF, creating \(H*W\) nodes in each stage\footnote{For instance, ResNet-18 consists of 4 stages [c2, c3, c4, c5] with [2, 2, 2, 2] blocks in each stage.}. 
% It is essential to model this spatial graph effectively to capture the spatial context, considering that spatial graphs differ across stages.

\begin{figure}%[!ht]%
\centering
\includegraphics[width=0.45\textwidth]{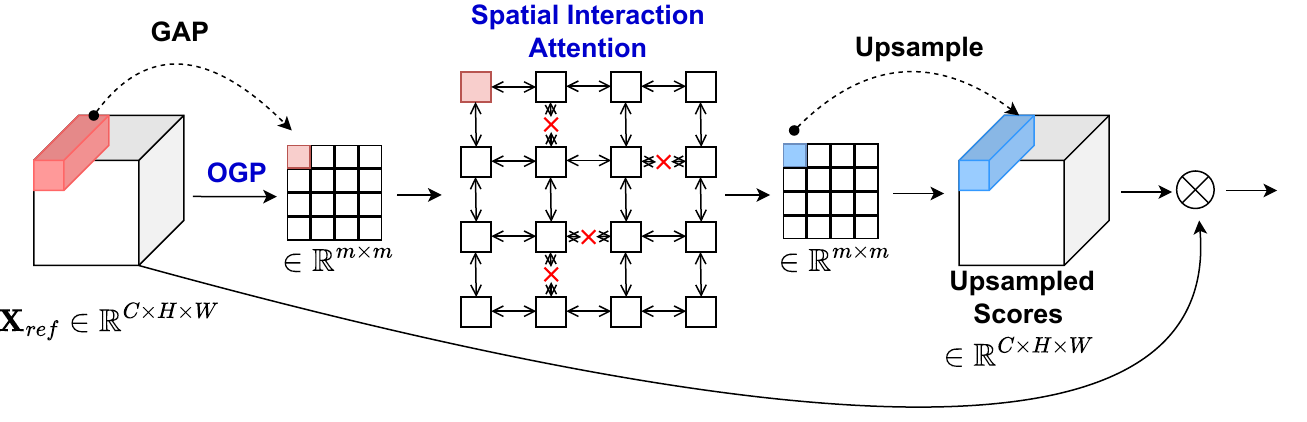}  \hfill
\caption{Spatial Interactional Attention Module.}
\label{figure:spatialattention}
\end{figure}

CBAM's spatial attention module follows SAMF and incorporates 1-strided convolution to generate pixel-wise spatial scores, combining both average and max pooling across channels. However, it adds extra computations to capture finer details, thus introducing redundancy in the early stages.
In order to effectively encapsulate spatial context with efficient computation, we present \textit{Output Guided Pooling} (OGP). OGP transforms the intermediate feature map \(\mathbb{R}^{C \times H \times W}\) into a fixed-size spatial map (\(\in \mathbb{R}^{(m \times m)}\)) using GAP along the channel dimension, where \(m\) corresponds to the dimension of the deepest spatial map (e.g., \(7 \times 7\) for ResNets). 
% Our model learns 'what' to attend and now needs to encode 'where' to focus spatially, which leads to the development of our Spatial Interactional Attention (SIA) module. 
% Inspired by \cite{zagoruyko2016paying}, who define a spatial attention map as a mapping \(\mathbb{R}^{C \times H \times W} \mapsto \mathbb{R}^{H \times W}\) through channel statistics, we base our formulation on this approach. 
% As \cite{islam2020much,zagoruyko2016paying} describe, CNNs extract fine-level details in early layers, mid-layers highlight discriminative features, and deep layers produce rich, category-specific representations.
% CBAM’s pixel-wise computation adds complexity to early layers. To address this, we propose Output Guided Pooling (OGP), mapping the intermediate feature map \(\mathbb{R}^{C \times H \times W}\) to a fixed-size \(\mathbb{R}^{k \times k}\) spatial attention map via GAP along the channel dimension. 
This fixed-size spatial map ensures consistency across multiple layers: in deeper layers, it captures pixel-wise details, while in earlier layers, it provides adaptive spatial information at a high level. 
% \cite{kayhan2020translation} show that convolutional layers encode absolute spatial locations, which OGP preserves effectively. 
We now construct a refined spatial graph \(G_{s}\) using the fixed-size spatial map (\(\in \mathbb{R}^{(m \times m)}\))(Fig. \ref{figure:spatialattention}).
This graph consists of \(m^2\) nodes, where each node connects to its adjacent neighbors: central nodes have four connections, edge nodes have three, and corner nodes have two, resulting in \(2m(m-1)\) edges.
\(\textbf{X}_{s} \in \mathbb{R}^{m^{2} \times 1}\) represents the initial node feature for the spatial nodes, obtained by applying OGP on \(\textbf{X}_{ref}\), which is derived from our CIA module. 
% Analogous to CIA, the computation of multi-head scaled dot product graph attention facilitates the spatial attention modeling by analyzing relative spatial significance (Eqs. \ref{eq1}, \ref{eq2} and \ref{eq3}) through attentional message passing among pixels in the refined spatial map. 
Analogous to CIA, we leverage multi-head scaled dot-product graph attention to model spatial attention through message passing between pixels in the fixed-size spatial map. 
This process includes averaging attention scores across all heads, updating node representations, and applying the sigmoid \(\sigma\) activation to obtain initial spatial scores \(\alpha_{init} \in \mathbb{R}^{m \times m}\) which are then upsampled using \texttt{torch.repeat\_interleave()} (Eq.~\ref{speq3}) to produce the final spatial scores \(\alpha_{s} \in \mathbb{R}^{C \times H \times W}\).
Modifying Eqs. \ref{eq1}, \ref{eq2} and \ref{eq3}, we obtain:
% (Following a similar approach as in Eqs. \ref{eq1}, \ref{eq2} and \ref{eq3} but utilizing \(\textbf{X}_{s}\) instead of \(\textbf{X}_{c}\), the final output will be \(\alpha_{s} \otimes \textbf{X}\)).
\begin{equation}
\textbf{A}_{i,j}^{h} = \text{softmax}\left(\frac{(\textbf{W}_{K}\textbf{X}_{s_{i}})^{T}\textbf{W}_{Q}\textbf{X}_{s_{j}}}{d_{K}}\right) , \, \forall j \in N_{i} 
\label{speq1}
\end{equation}
\begin{equation}
\textbf{A} = \frac{1}{H}\sum_{h=1}^{H} \textbf{A}^{h} \, \, ,  \, \, \textbf{X}_{s_{att}} = \textbf{A}\textbf{X}_{s}
\label{speq2}
\end{equation}
\begin{equation}
    \alpha_{init} = \sigma(\textbf{X}_{s_{att}}), \, \,
    \alpha_{s} = \text{upsample}(\alpha_{init}) 
\label{speq3}
\end{equation}
\begin{equation}
\textbf{X}_{final} = \textbf{X} + \alpha_{s} \otimes \textbf{X}_{ref}
\label{speq4}
\end{equation}
    where \(\textbf{X}_{final} \in \mathbb{R}^{C \times H \times W}\) represents the final output of STEAM, while other notations remain consistent.
% For graph representation learning, we use the same attention computation technique as in CAF (Eq \ref{eq1}, \ref{eq2} and \ref{eq3}).
% We calculate multi-head scaled dot product attention between nodes and their neighbors, average the results across all heads, multiply by the initial features, and apply a sigmoid activation to obtain the spatial scores. The initial node features come from our OGP (\(\mathbb{R}^{k \times k \times 1}\)). 
We note that the central submatrix (\(\mathbb{R}^{(m-2) \times (m-2)}\)) of our refined spatial map contains several nodes with a uniform degree of 4. This could lead to over-smoothing, as message passing among these highly interconnected nodes may result in nearly identical node representations, thereby degrading performance. To address this issue, we introduce an effective edge drop technique during model training, where one edge is randomly dropped for each node within this central submatrix. Empirical ablations in Table 5 of Appendix, demonstrate that this edge drop technique improves the model's performance. 
We name this entire mechanism as \textit{Spatial Interaction Attention} (SIA).
% To prevent GNNs from oversmoothing within the central submatrix (\(\mathbb{R}^{(k-2) \times (k-2)}\)) of the \(k \times k\) map, where multiple nodes have overlapping connections, we introduce an effective edge drop. During model training, one edge is randomly dropped for each node in the central submatrix. Empirical ablations show that this edge drop enhances performance. 
% Our SIA module offers key advantages: (i) It introduces structure-preserved spatial attention through OGP. (ii) The multi-head mechanism and random edge drop enhance node representation robustness (iii) It is computationally more efficient than CBAM’s spatial attention (iv) SIA is the first module to model relational attention within the spatial map. 

\subsection{STEAM: Squeeze and Transform Enhanced Attention Module} \label{steammethodology}
% - in this section we explain our detailed architecure
% - Inspired by ECA for channel attention, ICLR 2017 and Non-Local for spatial attention , finally with CBAM for combined , state -> what and where are both necessary. 
% - based on empirical analysis , we conclude that CA-SA order is correct. 
% - extensive ablations - we tried with every every, then with 1 per color block . - A ResNet typically has 4 stages (blockgroups) commonly referred to as [c2,c3,c4,c5] with strides [4,8,16,32] relative to input image, respectively. Stacks [c2, c3, c4, c5] consist of multiple bottleneck blocks with residual connections (Rs50 - [3,4,6,3] ..based on this we conslude floor(Ck // 3 ), why 3 , becoz of the basic least number of resblocks inside. reader can model it accordingly for their respective study. 
In this section, we introduce the final architecture.
% Traditional channel attention mechanisms depend on spatial representation maps for modeling their attention, whereas spatial attention mechanims depend on channel-wise global information for modeling their attention.
Traditional channel attention mechanisms leverage spatial representation maps, while spatial attention mechanisms draw on global channel information, highlighting the interdependence between the two.
% Driven by this insight, we integrate both our CIA and SIA modules to enhance channel and spatial modeling. 
Building on this insight, we integrate our CIA and SIA modules to enhance both channel and spatial modeling simultaneously.
Our empirical results from ablation studies (Table.~\ref{ablationtable}) reveal that sequential arrangement outperforms the parallel one, with channel-first order yielding slightly better results than spatial-first. In our experiments, we integrate STEAM with ResNet and ShuffleNet-V2 backbones.
In ResNet (stages [c2, c3, c4, c5]), we initially insert STEAM before each residual connection. For ShuffleNet-V2 (stages [c2, c3, c4]), we insert STEAM before each channel shuffle operation, as described by the authors in the Appendix of \cite{ma2018shufflenet}.
Based on extensive ablations, we propose an adaptive approach (detailed in the ablations) for determining the number of STEAM units per stage, resulting in: [1, 1, 1, 1] for ResNet18; [1, 1, 2, 1] for ResNet50; [1, 1, 6, 1] for ResNet101; and [1, 2, 1] for ShuffleNet-V2. STEAM units are added at the end of each stage, and if additional units are required, they are evenly distributed within that stage\footnote{Explained with diagrams in Appendix.}. This adaptive configuration, guided by experimental insights, delivers optimal results.
%the optimal number of STEAM units while also delivering the desired results effectively.
% In our quest for computational efficiency, we experimented with various STEAM arrangements. Initially, we placed STEAM before the last residual connection of each ResNet stage, leveraging stage-efficient representations. Inspired by MHSA block from \cite{srinivas2021bottleneck} , we also tested placing STEAM in the final three blocks of ResNet . We then develop an adaptive approach, determining the number of blocks in each stage based to be \(ceil(C / 4)\) where \(C\) denotes number of channels: [1, 1, 1, 1] for ResNet18, [1, 1, 2, 1] for ResNet50, [1, 1, 6, 1] for ResNet101, and [1, 2, 1] for ShuffleNetV2. This adaptive arrangement, grounded in experimental insights and computational efficiency, yielded the best results, demonstrating the effectiveness of minimal STEAM integration into ResNet and ShuffleNet stages as opposed to most of the existing SOTA modules
\subsection{Parameter and Computational Complexity}
In this section, we examine the parameters and computational complexity of STEAM. The parameters required for CIA and SIA are independent of the number of channels and spatial elements, as our attention modeling relies on feature vector representation, which remains consistent given a hidden model dimension \(d\). Each of CIA and SIA contributes \(4d\) parameters (since both the key and query operations contribute \(d\) for weights and biases each). As a result, a single STEAM unit introduces \(8d\) parameters. The total number of parameters is given by \(\sum_{s=1}^{S}(8d \times \lceil N_{bs} / 4 \rceil)\), where \(S\) signifies the total number of stages in the model and \(N_{bs}\) indicates the number of blocks in stage \(s\). For our experiments, we typically use \(d = 8\) \footnote{We have experimented with multiple values of \(d\) (refer to Table 7 in Appendix).}. 
%Consequently, for ResNet-50, the number of parameters added by incorporating all STEAM units is \(5*(8*8) = 320\).
As a result, incorporating all STEAM units in ResNet-50 adds  \(5*(8*8) = 320\) parameters.

% STEAM adds only 320 parameters and 3.57e-3 GFLOPs, yielding a \(2\%\) Top1-accuracy increase over standard ResNet-50, demonstrating its effectiveness in terms of parameters, computation, and performance.

\section{Experiments}
In this section, we evaluate STEAM for large-scale image classification with ImageNet \cite{deng2009imagenet} and for object detection and instance segmentation with MS COCO \cite{lin2014microsoft}. We reproduced SE, CBAM, ECA, and GCT, the leading SOTA modules, for fair comparisons using the same architecture, data augmentation, and optimization parameters. All the experiments were conducted with the PyTorch framework \cite{paszke2019pytorch} and the MMDetection toolkit \cite{chen2019mmdetection}. In Tables 2-5, bolded values denote the best performance.

\begin{table}
	%\small
	\footnotesize
	\centering 
	\setlength{\tabcolsep}{4pt}
%	\scalebox{0.9}{ %
		\begin{tabular}  {c|c|c|c|c|c}
			
			\hline
			 Ablation & Type & Ad. Params & Ad. GFLOPs & Top-1 & Top-5 \\ \hline
			\multirow{3}{*}{\centering{a}} & ca-sa & 320 & 3.57e-3 & \textbf{77.20} & \textbf{93.63} \\
			\multirow{3}{*}{} & sa-ca & 320 & 3.57e-3 & 76.94 & 93.46 \\
			\multirow{3}{*}{} & ca+sa & 320 & 3.57e-3 & 76.38 & 93.10 \\
            
			% +ECA & 34.5 & 56.4 & 36.4 \\
			% +GCT & 34.8 & 56.8 & 37.1 \\ 
			% +MCA-E & 35.3 & \textbf{57.6} & 37.3 \\ 
			% +MCA-S & \textbf{35.4} & 57.4&-- \\ 
			\hline
            \multirow{2}{*}{\centering{b}} & Degree-2 & 320 & 3.57e-3 & \textbf{77.20} & \textbf{93.63} \\
			\multirow{2}{*}{} & Degree-4 & 320 & 3.57e-3 & 76.78 & 93.27 \\
            \hline
			%\cline{1-7}
			\multirow{4}{*}{\centering{c}} & [*, *, *, *] & 1024 & 1.15e-2 & 76.87 & 93.38 \\
			\multirow{4}{*}{} & [0, 0, 0, *] & 192 & 8e-4 & 76.71 & 93.28\\
			\multirow{4}{*}{} &  [1, 1, 1, 1] & 256 & 3.13e-3 & 77.02 & 93.50\\
			\multirow{4}{*}{}   & [1, 1, 2, 1] & 320 & 3.57e-3 & \textbf{77.20} & \textbf{93.63}\\ 
			% +GCT     & 36.7 & 59.0 & \textbf{38.9}\\ 
			% +MCA-E  &36.6 & 59.0 &\textbf{38.9}\\ 
			% +MCA-S  & \textbf{36.8} & \textbf{59.5} & 38.8 \\ 
			\hline
            
		\end{tabular}
%	}
	\caption{Ablations results for three standard ablations denoted by a, b and c. Ad. Params and Ad. GFLOPs indicate additional parameters and GFLOPs introduced by our module for the respective ablation experiment.}
	\label{ablationtable}
\end{table}

\begin{table}[t]
\small
\centering
%\scalebox{0.85}{ %
%\setlength{\tabcolsep}{4pt}
\begin{tabular}{ l|ll|ll }

  \hline
  Methods & Params & GFLOPs & Top-1 & Top-5 \\ \hline
  %\hline
  ResNet-18 & 11.69M & 1.8253 & 69.75 & 89.07 \\
  +SE  & +87.04K & 1.8269 & 70.53 & 89.67 \\ 
  +CBAM    & +89.86K  & 1.8278 & 70.70 & 89.91 \\ 
  +ECA    & +0.04K & 1.8268 & 70.48 & 89.63 \\
  +GCT   & +0.01K  & 1.8268   & 71.06 & 89.96 \\  
  +STEAM (Ours)  & +0.25K  & 1.8261 & \textbf{71.36} & \textbf{90.10} \\ 
  \hline
  ResNet-50 & 25.56M & 4.1324 & 75.22 & 92.52 \\
  +SE  & +2.51M & 4.1460 & 76.61 & 93.18 \\ 
  +CBAM    & +2.53M  & 4.1500 & 76.94 & 93.46 \\ 
  +ECA    & +0.08K & 4.1436 & 76.88 & 93.38 \\ 
  +GCT   & +0.02K  & 4.1435 & 77.03 & 93.52 \\  
  +STEAM (Ours)  & +0.32K  & 4.1360 & \textbf{77.20} & \textbf{93.64} \\  
  \hline
  ShuffleNet-V2 & 2.28M & 0.1520 & 66.24 & 87.11 \\
  +SE  & +132k & 0.1547 & 66.92 & 87.65 \\ 
  +CBAM & +136k & 0.1561 & -- & -- \\ 
  +ECA    & +0.04K & 0.1542 & 66.87 & 87.63 \\
  +GCT   & +0.01K & 0.1542 & \textbf{67.21} & \textbf{87.90} \\  
  +STEAM (Ours)  & +0.25K & 0.1524 & 66.89 & 87.55 \\ 
  \hline

\end{tabular}
%}
\caption{Image classification results of the state-of-the-art channel attention blocks on ImageNet dataset.}

\label{tbl:table5}
\end{table}

\subsection{Implementation Details}
In order to assess STEAM’s performance on ImageNet, we evaluate it using three well-known CNN backbones: ResNet-18, ResNet-50 and ShuffleNet-V2 (with 1x scaling factor). All models were trained on 4 Nvidia RTX 4090 GPUs utilizing a mini-batch size of 256 (64 images per GPU). We apply identical training procedure across all models, including random cropping to a size of \(224 \times 224\) and random horizontal flipping. We use the \textit{Stochastic Gradient Descent} (SGD) optimizer with a momentum of 0.9 and a weight decay of 1e-4. All models were trained for 100 epochs with an initial learning rate of 0.1, reduced by a factor of 10 at \(30^{th}\), \(60^{th}\), and \(90^{th}\) epochs.
While evaluating the models on validation set, we begin by resizing the input image to \(256 \times 256\) followed by a center crop of \(224 \times 224\).
%While evaluating the models on validation set, we first resize the input image to a size of \(256 \times 256\) then take a central crop of size \(224 \times 224\) and normalize the images.
We report the Top-1 and Top-5 accuracy along with parameter count and GFLOPs to evaluate both storage and computational efficiency. 

Furthermore, we test the efficiency of our model in object detection and instance segmentation on MS COCO dataset, employing Faster R-CNN \cite{ren2015faster}, Mask R-CNN \cite{he2017mask}, and RetinaNet \cite{lin2017focal} detectors, using ResNet-50 with \textit{Feature Pyramid Network} (FPN) \cite{lin2017feature} as a backbone, initialized with weights from ImageNet classification. All models were trained using MMDetection toolkit on 4 Nvidia RTX 4090 GPUs with a mini-batch size of 8 (2 images per GPU). Pre-processing steps include resizing the shorter edge to 800 while constraining the larger edge to a size of 1333, randomly flipping, and normalizing the input image. SGD with a momentum of 0.9 and weight decay of 1e-4 was used as the optimizer. The initial learning rate was set to 0.02, decreasing by a factor of 10 at the \(8^{th}\) and \(11^{th}\) epochs, with a total of 12 epochs for training. We report \textit{average precision }(AP), \( \text{AP}_{0.5} \),  \( \text{AP}_{0.75} \),  \( \text{AP}_{S} \), \( \text{AP}_{M} \),  \( \text{AP}_{L} \) along with the number of parameters and GFLOPs to evaluate storage and computational efficiency.
\textbf{NOTE:} As mentioned in our methodology, we adhere to the same adaptive strategy when incorporating STEAM into ResNet and ShuffleNet-V2 models for all experiments, unless otherwise specified. However, we introduce SE, CBAM, ECA, and GCT modules after each residual connection in ResNet and prior to each channel shuffle operation in ShuffleNet-V2.

\begin{table*}[h]
	\small
	
	\centering
        % \renewcommand{\arraystretch}{1.15}
	% \setlength{\tabcolsep}{4pt}
	% \scalebox{0.6}{ %
		\begin{tabular}	{c|l|lc|cccccc}
			\hline
			Detector & Methods &  Params & GFLOPs & AP & $AP_{0.5}$ & $AP_{0.75}$ & $AP_{S}$ & $AP_{M}$ & $AP_{L}$ \\ \hline
			%\hline
			\multirow{6}{*}{\centering Faster-RCNN} & ResNet-50 & 41.53M & 207.07 & 36.2 & 57.5 & 39.2 & 20.7 & 39.8 & 46.1 \\
			\multirow{6}{*}{}& +SE (CVPR'18) & +2.51M & 207.19  & 36.9 & 58.4 & 39.9 & 21.7 & 40.6 & 47.2 \\
			\multirow{6}{*}{}& +CBAM (ECCV'18)  & +2.53M & 207.20  & 37.3 & 59.0 & 40.3 & 21.9 & 41.1 & 47.6\\ 
			\multirow{6}{*}{}& +ECA (CVPR'20) & +0.08K & 207.18 & 37.5 & 59.7 & 40.7 & 21.8 & 41.5 & 47.3 \\
			\multirow{6}{*}{}& +GCT (CVPR'21) & +0.02K & 207.18 & 37.8 & 60.1 & \textbf{41.2} & 22.3 & 41.8 & 47.8  \\  
			\multirow{6}{*}{}& +STEAM (Ours) & +0.32K & 207.07  & \textbf{38.1} & \textbf{60.3} & \textbf{41.2} & \textbf{22.5} & \textbf{42.2} & \textbf{48.3}\\ 
			
			\cline{2-10}
			% \multirow{8}{*}{} &ResNet-101 & 60.52M & 283.14   & 38.2 & 60.0 & 42.3 & 22.2 & 43.0 & 50.2 \\
			% \multirow{8}{*}{} & +SE (CVPR'18) & +4.74M & 283.33 & 38.5 & 60.7 & 42.2 & 22.5 & 43.7 & 51.0 \\
			% \multirow{8}{*}{} & +CBAM (ECCV'18) & +4.78M & 283.35 & 38.7 & 60.8 & 42.5 & 22.6 & 43.9 & 51.3 \\ 
			% \multirow{8}{*}{} & +ECA (CVPR'20) & +0.17K & 283.32 & 39.3 & 61.0 & 43.1 & 23.1 & 44.2 & 51.4 \\ 
			% \multirow{8}{*}{} & +GCT (CVPR'21) & +0.03K & 283.32 & 40.0 & 61.8 & 43.6 & 23.7 & 44.5 & 51.6\\ 
			% \multirow{8}{*}{} & +STEAM (Ours) & -- & -- & -- & -- & -- & -- & -- & --\\ 
			\hline
			\multirow{6}{*}{\centering Mask-RCNN} & ResNet-50 & 44.18M & 275.58 & 37.0 & 58.8 & 40.6 & 21.3 & 40.7 & 48.3 \\
			\multirow{6}{*}{}& +SE (CVPR'18) & +2.51M & 275.69  & 37.7 & 59.5 & 41.3 & 22.0 & 41.6 & 48.8 \\
			\multirow{6}{*}{}& +CBAM (ECCV'18) & +2.53M & 275.70 & 38.0 & 60.3 & 41.6 & 22.2 & 42.1 & 49.5\\
			\multirow{6}{*}{}& +ECA (CVPR'20) & +0.08K & 275.69  & 38.2 & 60.1 & 41.6 & 22.3 & 42.1 & 49.3 \\
			\multirow{6}{*}{}& +GCT (CVPR'21) & +0.02K & 275.69  & 38.7 & 60.6 & 42.1 & 22.8 & 42.7 & 50.0 \\
			\multirow{6}{*}{}& +STEAM (Ours) &+0.32K & 275.58  & \textbf{39.0} & \textbf{61.0} & \textbf{42.3} & \textbf{23.1} & \textbf{43.1} & \textbf{50.3} \\
			\cline{2-10}
			% \multirow{8}{*}{} &ResNet-101 & 63.17M & 351.65  & 39.2 & 58.8 & 43.0 & 22.6 & 44.0 & 51.2 \\
			% \multirow{8}{*}{} & +SE (CVPR'18) & +4.74M & 351.84  & 39.3 & 60.2 & 43.6 & 23.1 & 44.6 & 51.9 \\
			% \multirow{8}{*}{} & +CBAM (ECCV'18) & +4.78M & 351.86 & 39.7 & 60.7 & 43.8 & 23.2 & 44.8 & 52.3 \\ 
			% \multirow{8}{*}{} & +ECA (CVPR'20) & +0.17K & 351.84 & 40.1 & 61.2 & 44.1 & 23.4 & 45.1 & 53.1 \\ 
			% \multirow{8}{*}{} & +GCT (CVPR'21) & +0.03K & 351.84 & 40.7 & \textbf{61.6} & 44.5 & 23.7 & 45.5 & \textbf{53.7} \\ 
			% \multirow{6}{*}{} & +STEAM (Ours) & -- & -- & -- & -- & -- & -- & -- & --\\ 
			\hline
			\multirow{6}{*}{ \centering Retina-Net} & ResNet-50 & 37.74M & 239.32 & 34.9 & 55.0 & 38.1 & 20.1 & 39.3 & 46.1 \\
			\multirow{6}{*}{}& +SE (CVPR'18) & +2.51M & 239.43 & 35.2 & 55.9 & 38.8 & 21.1 & 40.1 & 47.3 \\
			\multirow{6}{*}{}& +CBAM (ECCV'18) & +2.53M & 239.44 & 35.6 & 56.3 & 39.0 & 21.3 & 40.7 & 47.6 \\
			\multirow{6}{*}{}& +ECA (CVPR'20) &  +0.08K & 239.43 & 35.9 & 56.4 & 39.3 & 21.3 & 41.0 & 47.8 \\
			\multirow{6}{*}{}& +GCT (CVPR'21) & +0.02K & 239.43 & 36.7 & 57.0 & 39.5 & 21.7 & 41.2 & 48.1 \\
			\multirow{6}{*}{}& +STEAM (Ours) &+0.32K & 239.32  & \textbf{37.1} &\textbf{ 57.4} & \textbf{39.6} &\textbf{ 21.9} & \textbf{41.5} & \textbf{48.3} \\ 
			\cline{2-10}
			% \multirow{8}{*}{} &ResNet-101 & 56.74M & 315.39 & 37.4 & 57.6 & 40.0 & 20.9 & 41.8 & 49.4 \\
			% \multirow{8}{*}{} & +SE (CVPR'18) & +4.74M & 315.58  & 37.7 & 57.8 & 41.0 & 21.4 & 42.3 & 50.5 \\
			% \multirow{8}{*}{} & +CBAM (ECCV'18) & +4.78M & 315.59 & 37.9 & 57.9 & 41.2 & 21.5 & 42.5 & 50.8 \\ 
			% \multirow{8}{*}{} & +ECA (CVPR'20) &  +0.17K & 315.57 & 38.2 & 58.5 & 41.2 & 21.3 & 42.4 & 50.4 \\ 
			% \multirow{8}{*}{} & +GCT (CVPR'21) & +0.03K & 315.57 & 38.6 & 59.0 & 41.5 & 21.7 & \textbf{43.0} & 50.8 \\ 
			% \multirow{8}{*}{} & +STEAM (Ours) & -- & -- & -- & -- & -- & -- & -- & --\\ 
			\hline
			
		\end{tabular}
	% }
	\caption{ Comparisons between different methods on the COCO val2017 set with object detection task.}
	\label{tbl:table4}
\end{table*}

\begin{table}
	%\small
	\footnotesize
	\centering 
	\setlength{\tabcolsep}{4pt}
%	\scalebox{0.9}{ %
		\begin{tabular}  { l|cccccc }
			
			\hline
			Methods & AP & $AP_{0.5}$ & $AP_{0.75}$ & $AP_{S}$ & $AP_{M}$ & $AP_{L}$ \\ \hline
			
			ResNet50 & 34.0 & 55.3 & 36.2 & 16.8 & 36.4 & 47.4 \\
			+SE  & 34.6 & 56.4 & 37.0 & 17.4 & 37.5 & 48.1  \\
			+CBAM  & 34.7 & 56.8 & 37.3 & 17.5 & 37.9 & 48.4 \\
			+ECA & 35.0 & 57.1 & 37.5 & 18.0 & 38.3 & 48.9 \\
			+GCT & 35.2 & \textbf{57.6} & 38.0 & 18.2 & 38.5 & 48.7 \\ 
			+STEAM (Ours) & \textbf{35.6} & 57.6 & 37.7 & \textbf{18.5} & \textbf{38.7} & \textbf{49.1}  \\
			\hline
			%\cline{1-7}
			% ResNet101  & 36.1 & 57.1 & 38.6 & 17.2 & 39.5 & 49.5 \\
			% +SE       & 36.5 & 57.9 & 39.2 & 17.8 & 40.3 & 50.1 \\
			% +CBAM  & 36.8 & 58.1 & 39.6 & 18.0 & 40.6 & 50.6 \\
			% +ECA   & 37.2 & 58.5 & 39.5 & 18.3 & 41.2 & 51.0 \\ 
			% +GCT     & \textbf{37.5} & 59.0 & \textbf{39.8} & 18.4 & 41.0 & 51.3 \\ 
			% +STEAM (Ours)  & -- & -- & -- & -- & -- & --  \\ 
			% \hline
		\end{tabular}
%	}
	\caption{Instance Segmentation using the state-of-the-art channel attention blocks on COCO val2017 set. }
	\label{tbl:table3}
\end{table}

\subsection{Ablation Studies}
We conduct extensive ablation studies to demonstrate the effectiveness of STEAM's design choice. These ablations employ ResNet-50 integrated with STEAM for image classification on the ImageNet dataset. Table \ref{ablationtable} shows the results of three ablation experiments: the first evaluates the arrangement of CIA and SIA modules, the second identifies the suitable neighborhood size for CIA, and the third examines the best placement and number of STEAM units within a deep CNN. These ablations are denoted by Ablation-a, Ablation-b, and Ablation-c, respectively.
%\subsection{Arrangement of CIA and SIA modules}

\noindent
\textbf{Arrangement of CIA and SIA modules} Here we perform three experiments to identify the optimum ordering of CIA and SIA modules. We employ a parallel arrangement of CIA and SIA modules (denoted as ``ca+sa") along with two sequential arrangements, CIA-first and SIA-first (denoted by ``ca-sa" and ``sa-ca" respectively) to evaluate the impact of different module arrangements. From Table \ref{ablationtable}, we can infer that the sequential arrangement outperforms the parallel one and a CIA-first order within the sequential arrangement slightly surpasses the SIA-first order.

\noindent
% \subsection{Choosing the Neighborhood Size for CIA} 
\textbf{Choosing the Neighbourhood Size for CIA} We experiment with two different node neighborhood sizes for our channel graph \(G_{c}\). First, we connect each channel's node to its direct 1-hop neighbors, giving each node 2 neighbors. We extend the neighborhood size to four by linking each node to its two preceding and two succeeding nodes (similar to ECA when its kernel size equals 5). Observe that in both scenarios, a cyclic graph structure is formed, with each node having a degree of either two or four. As shown in Table \ref{ablationtable}, configuring each channel node with degree 2 yields better results than degree 4.

% \subsection{Arrangement of STEAM in a deep CNN}
\noindent
\textbf{Arrangement of STEAM in a deep CNN} Inspired by current SOTA architectures, we initially integrated STEAM after every block within each stage of ResNet-50, which contains [3, 4, 6, 3] blocks in stages [c2, c3, c4, c5], respectively, resulting in a total of 16 STEAM units. However, empirical findings indicate that this approach produced suboptimal performance. Following the approach described in \cite{srinivas2021bottleneck}, we added STEAM only in the final stage of ResNet-50, reducing it to 3 STEAM units. Furthermore, we tested the addition of a single STEAM unit after each stage. As shown in Table \ref{ablationtable}, adding STEAM after every stage yielded better results compared to adding STEAM only in the final stage. Consequently, we adopted a dynamic strategy to tailor the optimal number of STEAM units per stage based on the number of blocks within that stage. Formally, the number of STEAM units per stage is given by \(\lceil(N_{bs} / 4 \rceil)\), where \(N_{bs}\) denotes the number of blocks in a stage. Specifically, for ResNet-50, the number of STEAM units per stage is [1, 1, 2, 1].

\subsection{Image Classification on ImageNet-1K}
This section presents a comparison of our model against SOTA counterparts. We selected three popular backbone models: ResNet-18, ResNet-50, and ShuffleNet-V2. The empirical results are displayed in Table~\ref{tbl:table5}. 

\textbf{ResNet} We initially assess the performance of STEAM using ResNet-18, and ResNet-50 as backbone models. Table~\ref{tbl:table5} clearly shows that STEAM surpasses all its counterparts across various ResNet models. Notably, in ResNet-18, STEAM achieves a 0.30\% increase in Top-1 accuracy over GCT, which already outperforms other modules. For ResNet-50, STEAM surpasses the former top module, GCT, by 0.17\% in Top-1 accuracy, respectively. 
Overall, STEAM provides an accuracy boost of 1.61\% and 1.98\% over the standard ResNet-18 and ResNet-50 models, respectively.

\textbf{ShuffleNet-V2} We evaluated STEAM on a lightweight CNN architecture\footnote{tested with more lightweight architectures in our Appendix.}, ShuffleNet-V2 as the backbone. As illustrated in Table~\ref{tbl:table5}, STEAM surpasses all other modules, obtaining 0.27\% Top-1 accuracy gain, compared to the second-best module, GCT. In general, STEAM outperforms all prior SOTA modules while maintaining comparable parameter counts and computational complexity.

\subsection{Object Detection on MS COCO}
% We carry out object detection experiments on the MS-COCO dataset employing Faster-RCNN, Mask-RCNN, and RetinaNet as detectors. For the backbone models, we use both ResNet-50 and ResNet-101. We compare the performance of our approach with modules such as SE, CBAM, ECA, GCT, and MCA. As shown in Table~\ref{tbl:table4}, our method outperforms all previous SOTA approaches in object detection.
We conduct object detection experiments on the MS-COCO dataset using Faster R-CNN, Mask R-CNN, and RetinaNet detectors with ResNet-50 as the backbone. We compare our approach with SE, CBAM, ECA, and GCT modules. As shown in Table~\ref{tbl:table4}, STEAM consistently outperforms these SOTA methods in object detection.

% In experiments using Faster-RCNN with ResNet-50 as the backbone, STEAM demonstrates superior performance compared to the baseline model, SE, CBAM, ECA, GCT, and MCA methods, showing an increase in AP by 1.9\%, 1.2\%, 0.8\%, 0.6\%, 0.3\%, and 0.5\% respectively. Similarly, when ResNet-101 is employed as the backbone, STEAM achieves higher AP than the baseline model, SE, CBAM, ECA, GCT, and MCA methods, with improvements of 2.0\%, 1.7\%, 1.5\%, 0.9\%, 0.4\%, and 0.8\% respectively. 
For Faster-RCNN with ResNet-50, STEAM improves AP by \(1.9\%\) over the baseline, with gains of 1.2\% over SE, 0.8\% over CBAM, 0.6\% over ECA, and 0.3\% over GCT. 
% With ResNet-101, STEAM achieves a 2.0\% increase over the baseline in AP, outperforming SE by 1.7\%, CBAM by 1.5\%, ECA by 0.9\%, GCT by 0.2\%, and MCA by 0.8\%.
% In Mask-RCNN experiments utilizing ResNet-50 as the backbone, STEAM demonstrates superior performance compared to the baseline model, SE, CBAM, ECA, GCT, and MCA methods, achieving higher AP by 2.0\%, 1.4\%, 1.1\%, 0.9\%, 0.3\%, and 0.7\% respectively. Furthermore, with ResNet-101 as the backbone, STEAM also attains higher AP over the baseline model, SE, CBAM, ECA, GCT, and MCA methods with improvements of 1.8\%, 1.7\%, 1.3\%, 0.9\%, 0.4\%, and 0.5\% respectively.
In Mask-RCNN experiments with ResNet-50 as the backbone, STEAM outperforms the baselines SE, CBAM, ECA, and GCT, achieving AP gains of 2.0\%, 1.3\%, 1.0\%, 0.8\%, and 0.3\%, respectively. 
% With ResNet-101, STEAM also surpasses these methods, achieving AP improvements of 1.8\%, 1.7\%, 1.3\%, 0.9\%, 0.3\%, and 0.5\%.
% In RetinaNet experiments utilizing ResNet-50 as the backbone, STEAM demonstrates enhanced performance compared to the baseline model and the methods SE, CBAM, ECA, GCT, and MCA, with respective increases in AP of 2.2\%, 1.9\%, 1.5\%, 1.2\%, 0.4\%, and 1.0\%. Subsequently, when using ResNet-101 as the backbone, STEAM achieves higher AP in comparison to the baseline model and the SE, CBAM, ECA, GCT, and MCA methods, with improvements of 1.6\%, 1.3\%, 1.1\%, 0.8\%, 0.4\%, and 0.8\%, respectively.
In RetinaNet experiments with ResNet-50, STEAM outperforms the baselines SE, CBAM, ECA, and GCT, achieving AP gains of 2.2\%, 1.9\%, 1.5\%, 1.2\%, and 0.4\% respectively. 
% With ResNet-101, STEAM also surpasses these methods, achieving AP improvements of 1.6\%, 1.3\%, 1.1\%, 0.8\%, 0.4\%, and 0.8\%.

\subsection{Instance Segmentation on MS-COCO dataset}
Finally, we perform instance segmentation experiments on the MS-COCO dataset using Mask R-CNN as the detector and ResNet-50 as the backbone. As presented in Table~\ref{tbl:table3}, our proposed method, STEAM, significantly exceeds the current state-of-the-art modules. Specifically, with ResNet-50 as the backbone, STEAM surpasses the baseline, SE, CBAM, ECA, and GCT methods by improvements of 1.6\%, 1.0\%, 0.9\%, 0.6\%, and 0.4\% AP, respectively. 
% Likewise, employing ResNet-101 as the backbone model, STEAM demonstrates superior AP performance over the baseline model, SE, CBAM, ECA, and MCA methods with increases of 1.3\%, 0.9\%, 0.6\%, 0.2\%, and 0.3\% respectively. 
% Notably, in this experiment, GCT surpasses STEAM by a marginally higher 0.1\% AP score.
Overall, these results highlight the effectiveness and generalization capabilities of our proposed STEAM method.

\section{Conclusion}
In this paper, we focus on learning effective channel and spatial attention for deep CNNs using a graphical approach by constructing channel and spatial graphs \(G_{c}\) and \(G_{s}\), supported by various motivations and references. 
% We employ multi-head scaled dot-product graph attention to capture diverse inter-channel and inter-spatial relationships. 
% To optimally model spatial attention, we introduced Output Guided Pooling. 
% We employ multi-head scaled dot-product graph attention to capture diverse inter-channel (CIA) and inter-spatial (SIA) relationships, and introduce Output Guided Pooling to optimally model spatial attention.
We propose CIA and SIA modules, which employ multi-head scaled dot-product graph attention to capture diverse inter-channel and inter-spatial relationships respectively and introduce OGP to model spatial attention optimally.
% STEAM, our proposed method, efficiently models both channel and spatial attention with low parameter and computational costs. 
% STEAM can be seamlessly integrated with multiple backbones such as ResNets, ShuffleNets
Extensive ablations support the construction of our proposed method STEAM, which efficiently models both channel and spatial attention with low parameters and computational costs. Moreover, STEAM can be seamlessly integrated with various backbones, including ResNets and lightweight models such as ShuffleNet-V2.
Experimental results demonstrate that STEAM outperforms previous SOTA methods in classification, object detection and instance segmentation tasks.
In future research, we intend to investigate several promising directions: 
% (1) Integrating positional encoding into spatial attention mechanism to better capture positional data from spatial contexts;
(1) Integrating positional encoding into spatial attention to better capture spatial context;
% (2) Improving correlation graph learning by employing edge modeling techniques
(2) Enhancing correlation graph learning through edge modeling techniques.

\bibliography{aaai25}

@inproceedings{he2016deep,
  title={Deep residual learning for image recognition},
  author={He, Kaiming and Zhang, Xiangyu and Ren, Shaoqing and Sun, Jian},
  booktitle={Proceedings of the IEEE conference on computer vision and pattern recognition},
  pages={770--778},
  year={2016}
}

@article{hu2018gather,
  title={Gather-excite: Exploiting feature context in convolutional neural networks},
  author={Hu, Jie and Shen, Li and Albanie, Samuel and Sun, Gang and Vedaldi, Andrea},
  journal={Advances in neural information processing systems},
  volume={31},
  year={2018}
}

@inproceedings{hu2018squeeze,
  title={Squeeze-and-excitation networks},
  author={Hu, Jie and Shen, Li and Sun, Gang},
  booktitle={Proceedings of the IEEE conference on computer vision and pattern recognition},
  pages={7132--7141},
  year={2018}
}

@inproceedings{wang2020eca,
  title={ECA-Net: Efficient channel attention for deep convolutional neural networks},
  author={Wang, Qilong and Wu, Banggu and Zhu, Pengfei and Li, Peihua and Zuo, Wangmeng and Hu, Qinghua},
  booktitle={Proceedings of the IEEE/CVF conference on computer vision and pattern recognition},
  pages={11534--11542},
  year={2020}
}

@article{park2018bam,
  title={Bam: Bottleneck attention module},
  author={Park, Jongchan and Woo, Sanghyun and Lee, Joon-Young and Kweon, In So},
  journal={arXiv preprint arXiv:1807.06514},
  year={2018},
}

@inproceedings{li2019selective,
  title={Selective kernel networks},
  author={Li, Xiang and Wang, Wenhai and Hu, Xiaolin and Yang, Jian},
  booktitle={Proceedings of the IEEE/CVF conference on computer vision and pattern recognition},
  pages={510--519},
  year={2019}
}

@inproceedings{wang2018non,
  title={Non-local neural networks},
  author={Wang, Xiaolong and Girshick, Ross and Gupta, Abhinav and He, Kaiming},
  booktitle={Proceedings of the IEEE conference on computer vision and pattern recognition},
  pages={7794--7803},
  year={2018}
}

@inproceedings{cao2019gcnet,
  title={Gcnet: Non-local networks meet squeeze-excitation networks and beyond},
  author={Cao, Yue and Xu, Jiarui and Lin, Stephen and Wei, Fangyun and Hu, Han},
  booktitle={Proceedings of the IEEE/CVF international conference on computer vision workshops},
  pages={0--0},
  year={2019}
}

@inproceedings{ruan2021gaussian,
  title={Gaussian context transformer},
  author={Ruan, Dongsheng and Wang, Daiyin and Zheng, Yuan and Zheng, Nenggan and Zheng, Min},
  booktitle={Proceedings of the IEEE/CVF Conference on Computer Vision and Pattern Recognition},
  pages={15129--15138},
  year={2021},
}

@inproceedings{lee2019srm,
  title={Srm: A style-based recalibration module for convolutional neural networks},
  author={Lee, HyunJae and Kim, Hyo-Eun and Nam, Hyeonseob},
  booktitle={Proceedings of the IEEE/CVF International conference on computer vision},
  pages={1854--1862},
  year={2019}
}

@article{jaderberg2015spatial,
  title={Spatial transformer networks},
  author={Jaderberg, Max and Simonyan, Karen and Zisserman, Andrew and others},
  journal={Advances in neural information processing systems},
  volume={28},
  year={2015}
}

@inproceedings{wang2017residual,
  title={Residual attention network for image classification},
  author={Wang, Fei and Jiang, Mengqing and Qian, Chen and Yang, Shuo and Li, Cheng and Zhang, Honggang and Wang, Xiaogang and Tang, Xiaoou},
  booktitle={Proceedings of the IEEE conference on computer vision and pattern recognition},
  pages={3156--3164},
  year={2017}
}

@inproceedings{zhang2021sa,
  title={Sa-net: Shuffle attention for deep convolutional neural networks},
  author={Zhang, Qing-Long and Yang, Yu-Bin},
  booktitle={ICASSP 2021-2021 IEEE International Conference on Acoustics, Speech and Signal Processing (ICASSP)},
  pages={2235--2239},
  year={2021},
  organization={IEEE}
}

@inproceedings{woo2018cbam,
  title={Cbam: Convolutional block attention module},
  author={Woo, Sanghyun and Park, Jongchan and Lee, Joon-Young and Kweon, In So},
  booktitle={Proceedings of the European conference on computer vision (ECCV)},
  pages={3--19},
  year={2018}
}

@article{bahdanau2014neural,
  author={Bahdanau, Dzmitry and Cho, Kyunghyun and Bengio, Yoshua},
  title={Neural machine translation by jointly learning to align and translate},
  year={2014},
  journal={arXiv preprint arXiv:1409.0473}, 
}

@article{chen20182,
  author={Chen, Yunpeng and Kalantidis, Yannis and Li, Jianshu and Yan, Shuicheng and Feng, Jiashi},
  title={A\^{} 2-nets: Double attention networks},
  volume={31},
  year={2018},
  journal={Advances in neural information processing systems},
}

@article{vaswani2017attention,
  title={Attention is all you need},
  author={Vaswani, Ashish and Shazeer, Noam and Parmar, Niki and Uszkoreit, Jakob and Jones, Llion and Gomez, Aidan N and Kaiser, {\L}ukasz and Polosukhin, Illia},
  journal={Advances in neural information processing systems},
  volume={30},
  year={2017}
}

@article{scarselli2008graph,
  title={The graph neural network model},
  author={Scarselli, Franco and Gori, Marco and Tsoi, Ah Chung and Hagenbuchner, Markus and Monfardini, Gabriele},
  journal={IEEE transactions on neural networks},
  volume={20},
  number={1},
  pages={61--80},
  year={2008},
  publisher={IEEE}
}

@inproceedings{gilmer2017neural,
  title={Neural message passing for quantum chemistry},
  author={Gilmer, Justin and Schoenholz, Samuel S and Riley, Patrick F and Vinyals, Oriol and Dahl, George E},
  booktitle={International conference on machine learning},
  pages={1263--1272},
  year={2017},
  organization={PMLR}
}

@article{velivckovic2017graph,
  title={Graph attention networks},
  author={Veli{\v{c}}kovi{\'c}, Petar and Cucurull, Guillem and Casanova, Arantxa and Romero, Adriana and Lio, Pietro and Bengio, Yoshua},
  journal={arXiv preprint arXiv:1710.10903},
  year={2017}
}

@article{brody2021attentive,
  title={How attentive are graph attention networks?},
  author={Brody, Shaked and Alon, Uri and Yahav, Eran},
  journal={arXiv preprint arXiv:2105.14491},
  year={2021}
}

@inproceedings{gao2019global,
  title={Global second-order pooling convolutional networks},
  author={Gao, Zilin and Xie, Jiangtao and Wang, Qilong and Li, Peihua},
  booktitle={Proceedings of the IEEE/CVF Conference on computer vision and pattern recognition},
  pages={3024--3033},
  year={2019}
}

@article{zagoruyko2016paying,
  title={Paying more attention to attention: Improving the performance of convolutional neural networks via attention transfer},
  author={Zagoruyko, Sergey and Komodakis, Nikos},
  journal={arXiv preprint arXiv:1612.03928},
  year={2016}
}

@article{roy2018recalibrating,
  title={Recalibrating fully convolutional networks with spatial and channel “squeeze and excitation” blocks},
  author={Roy, Abhijit Guha and Navab, Nassir and Wachinger, Christian},
  journal={IEEE transactions on medical imaging},
  volume={38},
  number={2},
  pages={540--549},
  year={2018},
  publisher={IEEE}
}

@inproceedings{zhang2018shufflenet,
  title={Shufflenet: An extremely efficient convolutional neural network for mobile devices},
  author={Zhang, Xiangyu and Zhou, Xinyu and Lin, Mengxiao and Sun, Jian},
  booktitle={Proceedings of the IEEE conference on computer vision and pattern recognition},
  pages={6848--6856},
  year={2018}
}

@inproceedings{ma2018shufflenet,
  title={Shufflenet v2: Practical guidelines for efficient cnn architecture design},
  author={Ma, Ningning and Zhang, Xiangyu and Zheng, Hai-Tao and Sun, Jian},
  booktitle={Proceedings of the European conference on computer vision (ECCV)},
  pages={116--131},
  year={2018}
}

@article{islam2020much,
  title={How much position information do convolutional neural networks encode?},
  author={Islam, Md Amirul and Jia, Sen and Bruce, Neil DB},
  journal={arXiv preprint arXiv:2001.08248},
  year={2020}
}

@article{kipf2016semi,
  title={Semi-supervised classification with graph convolutional networks},
  author={Kipf, Thomas N and Welling, Max},
  journal={arXiv preprint arXiv:1609.02907},
  year={2016}
}

@article{hamilton2017inductive,
  title={Inductive representation learning on large graphs},
  author={Hamilton, Will and Ying, Zhitao and Leskovec, Jure},
  journal={Advances in neural information processing systems},
  volume={30},
  year={2017}
}

@article{dwivedi2020generalization,
  title={A generalization of transformer networks to graphs},
  author={Dwivedi, Vijay Prakash and Bresson, Xavier},
  journal={arXiv preprint arXiv:2012.09699},
  year={2020}
}

@article{battaglia2018relational,
  title={Relational inductive biases, deep learning, and graph networks},
  author={Battaglia, Peter W and Hamrick, Jessica B and Bapst, Victor and Sanchez-Gonzalez, Alvaro and Zambaldi, Vinicius and Malinowski, Mateusz and Tacchetti, Andrea and Raposo, David and Santoro, Adam and Faulkner, Ryan and others},
  journal={arXiv preprint arXiv:1806.01261},
  year={2018}
}

@article{kreuzer2021rethinking,
  title={Rethinking graph transformers with spectral attention},
  author={Kreuzer, Devin and Beaini, Dominique and Hamilton, Will and L{\'e}tourneau, Vincent and Tossou, Prudencio},
  journal={Advances in Neural Information Processing Systems},
  volume={34},
  pages={21618--21629},
  year={2021}
}

@article{rampavsek2022recipe,
  title={Recipe for a general, powerful, scalable graph transformer},
  author={Ramp{\'a}{\v{s}}ek, Ladislav and Galkin, Michael and Dwivedi, Vijay Prakash and Luu, Anh Tuan and Wolf, Guy and Beaini, Dominique},
  journal={Advances in Neural Information Processing Systems},
  volume={35},
  pages={14501--14515},
  year={2022}
}

@article{han2022vision,
  title={Vision gnn: An image is worth graph of nodes},
  author={Han, Kai and Wang, Yunhe and Guo, Jianyuan and Tang, Yehui and Wu, Enhua},
  journal={Advances in neural information processing systems},
  volume={35},
  pages={8291--8303},
  year={2022}
}

@inproceedings{aflalo2023deepcut,
  title={Deepcut: Unsupervised segmentation using graph neural networks clustering},
  author={Aflalo, Amit and Bagon, Shai and Kashti, Tamar and Eldar, Yonina},
  booktitle={Proceedings of the IEEE/CVF International Conference on Computer Vision},
  pages={32--41},
  year={2023}
}

@article{jiao2022graph,
  title={Graph representation learning meets computer vision: A survey},
  author={Jiao, Licheng and Chen, Jie and Liu, Fang and Yang, Shuyuan and You, Chao and Liu, Xu and Li, Lingling and Hou, Biao},
  journal={IEEE Transactions on Artificial Intelligence},
  volume={4},
  number={1},
  pages={2--22},
  year={2022},
  publisher={IEEE}
}

@article{dosovitskiy2020image,
  title={An image is worth 16x16 words: Transformers for image recognition at scale},
  author={Dosovitskiy, Alexey and Beyer, Lucas and Kolesnikov, Alexander and Weissenborn, Dirk and Zhai, Xiaohua and Unterthiner, Thomas and Dehghani, Mostafa and Minderer, Matthias and Heigold, Georg and Gelly, Sylvain and others},
  journal={arXiv preprint arXiv:2010.11929},
  year={2020}
}

@inproceedings{srinivas2021bottleneck,
  title={Bottleneck transformers for visual recognition},
  author={Srinivas, Aravind and Lin, Tsung-Yi and Parmar, Niki and Shlens, Jonathon and Abbeel, Pieter and Vaswani, Ashish},
  booktitle={Proceedings of the IEEE/CVF conference on computer vision and pattern recognition},
  pages={16519--16529},
  year={2021}
}

@article{nt2019revisiting,
  title={Revisiting graph neural networks: All we have is low-pass filters},
  author={Nt, Hoang and Maehara, Takanori},
  journal={arXiv preprint arXiv:1905.09550},
  year={2019}
}

@article{oono2019graph,
  title={Graph neural networks exponentially lose expressive power for node classification},
  author={Oono, Kenta and Suzuki, Taiji},
  journal={arXiv preprint arXiv:1905.10947},
  year={2019}
}

@article{roweis2000nonlinear,
  title={Nonlinear dimensionality reduction by locally linear embedding},
  author={Roweis, Sam T and Saul, Lawrence K},
  journal={science},
  volume={290},
  number={5500},
  pages={2323--2326},
  year={2000},
  publisher={American Association for the Advancement of Science}
}

@inproceedings{deng2009imagenet,
  title={Imagenet: A large-scale hierarchical image database},
  author={Deng, Jia and Dong, Wei and Socher, Richard and Li, Li-Jia and Li, Kai and Fei-Fei, Li},
  booktitle={2009 IEEE conference on computer vision and pattern recognition},
  pages={248--255},
  year={2009},
  organization={Ieee}
}

@article{paszke2019pytorch,
  title={Pytorch: An imperative style, high-performance deep learning library},
  author={Paszke, Adam and Gross, Sam and Massa, Francisco and Lerer, Adam and Bradbury, James and Chanan, Gregory and Killeen, Trevor and Lin, Zeming and Gimelshein, Natalia and Antiga, Luca and others},
  journal={Advances in neural information processing systems},
  volume={32},
  year={2019}
}

@inproceedings{lin2014microsoft,
  title={Microsoft coco: Common objects in context},
  author={Lin, Tsung-Yi and Maire, Michael and Belongie, Serge and Hays, James and Perona, Pietro and Ramanan, Deva and Doll{\'a}r, Piotr and Zitnick, C Lawrence},
  booktitle={Computer Vision--ECCV 2014: 13th European Conference, Zurich, Switzerland, September 6-12, 2014, Proceedings, Part V 13},
  pages={740--755},
  year={2014},
  organization={Springer}
}

@article{chen2019mmdetection,
  title={MMDetection: Open mmlab detection toolbox and benchmark},
  author={Chen, Kai and Wang, Jiaqi and Pang, Jiangmiao and Cao, Yuhang and Xiong, Yu and Li, Xiaoxiao and Sun, Shuyang and Feng, Wansen and Liu, Ziwei and Xu, Jiarui and others},
  journal={arXiv preprint arXiv:1906.07155},
  year={2019}
}

@article{ren2015faster,
  title={Faster r-cnn: Towards real-time object detection with region proposal networks},
  author={Ren, Shaoqing and He, Kaiming and Girshick, Ross and Sun, Jian},
  journal={Advances in neural information processing systems},
  volume={28},
  year={2015}
}

@inproceedings{he2017mask,
  title={Mask r-cnn},
  author={He, Kaiming and Gkioxari, Georgia and Doll{\'a}r, Piotr and Girshick, Ross},
  booktitle={Proceedings of the IEEE international conference on computer vision},
  pages={2961--2969},
  year={2017}
}

@inproceedings{lin2017focal,
  title={Focal loss for dense object detection},
  author={Lin, Tsung-Yi and Goyal, Priya and Girshick, Ross and He, Kaiming and Doll{\'a}r, Piotr},
  booktitle={Proceedings of the IEEE international conference on computer vision},
  pages={2980--2988},
  year={2017}
}

@inproceedings{xie2017aggregated,
  title     = {Aggregated residual transformations for deep neural networks},
  author    = {Xie, Saining and Girshick, Ross and Doll{\'a}r, Piotr and Tu, Zhuowen and He, Kaiming},
  booktitle = {Proceedings of the IEEE conference on computer vision and pattern recognition},
  pages     = {1492--1500},
  year      = {2017}
}

@inproceedings{lin2017feature,
  title={Feature pyramid networks for object detection},
  author={Lin, Tsung-Yi and Doll{\'a}r, Piotr and Girshick, Ross and He, Kaiming and Hariharan, Bharath and Belongie, Serge},
  booktitle={Proceedings of the IEEE conference on computer vision and pattern recognition},
  pages={2117--2125},
  year={2017}
}

@article{krizhevsky2012imagenet,
  title={Imagenet classification with deep convolutional neural networks},
  author={Krizhevsky, Alex and Sutskever, Ilya and Hinton, Geoffrey E},
  journal={Advances in neural information processing systems},
  volume={25},
  year={2012}
}

@inproceedings{wang2020understanding,
  title={Understanding contrastive representation learning through alignment and uniformity on the hypersphere},
  author={Wang, Tongzhou and Isola, Phillip},
  booktitle={International conference on machine learning},
  pages={9929--9939},
  year={2020},
  organization={PMLR}
}

\clearpage
\section{Appendix}
% \subsection{Note to reviewers}
% % In this section we discuss the general configuration of our proposed method, STEAM. Table \ref{tab:config-table} lists the general configuration of each of the CIA and  SIA modules. We also introduce a non-linearity between CIA and SIA which is fixed to Tanh.
% \textbf{In Eq. 1 and Eq. 4 of our main paper, we mentioned \(\textbf{W}_{K}, \textbf{W}_{Q} \in \mathbb{R}^{d_{K} \times d}\). The intended dimensions are \(\textbf{W}_{K}, \textbf{W}_{Q} \in \mathbb{R}^{d_{K} \times d_{in}}\), where \(d_{in} = 1\) in our case. This was an oversight, as \(d\) represents the hidden dimension used in our multi-head graph attention, while \(d_{in}\) denotes the input dimension and is always equal to 1 in both CIA and SIA as \(\textbf{X}_{c} \in \mathbb{R}^{C \times 1}\) and \(\textbf{X}_{s} \in \mathbb{R}^{m^{2} \times 1}\). Additionally, in Fig. 1, by \(\textbf{Q}\) and \(\textbf{K}\) we mean \(\textbf{W}_{Q}\) and \(\textbf{W}_{K}\),itself respectively, with \(\textbf{Q}, \textbf{W} \in \mathbb{R}^{d_{K} \times d_{in}}\).
% % where \(d_{in}\) is always equal to 1 as \(\textbf{X}_{c} \in \mathbb{R}^{C \times 1}\) and \(\textbf{X}_{s} \in \mathbb{R}^{m^{2} \times 1}\)
% % . 
% We apologize for any confusion caused by this oversight.}

% \textbf{We maintain a consistent STEAM configuration across all ablations presented in both the main paper and the Appendix. Only the specific parameters being evaluated in the ablations are modified.}

\section{STEAM Configuration}
\begin{figure}%[!ht]%
\centering
\includegraphics[width=0.25\textwidth]{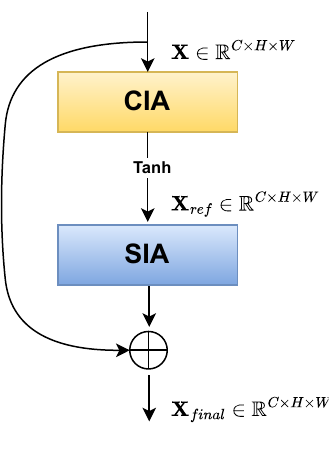}  \hfill
\caption{Overview of STEAM}
\label{figure:stream}
\end{figure}

In this section, we discuss the overall configuration of our proposed method, STEAM. Table \ref{tab:config-table} provides the configuration details for both the CIA and SIA modules. Additionally, we introduce a fixed \(\tanh\) non-linearity between the CIA and SIA modules.
The value of \(d_{K}\) (which determines the dimension for \(\textbf{W}_{K}\) and \(\textbf{W}_{Q}\) referenced in Eqs. 1 and 4) is not explicitly stated, but is often defined as \(d / H\), which equals 2 for both CIA and SIA, where \(H\) denotes the total number of heads. An overview of STEAM is shown in Fig.~\ref{figure:stream}

\renewcommand\arraystretch{1.3}
\begin{table*}[h]
\centering
\begin{tabular}{c|c|c}
\hline
\textbf{Parameter} & \textbf{Channel Interaction Attention (CIA)} & \textbf{Spatial Interaction Attention (SIA)} \\ \hline
Pooling mechanism & GAP & OGP with output size fixed to \(7 \times 7\) \\ \hline
Node degree & Fixed to 2 for each node & 2 - corner, 3 - edge and 4 - central nodes \\ \hline
\(d\) & 8 & 8 \\ \hline
\(H\) & 4 & 4 \\ \hline
Edge drop used & No & Yes (only for central submatrix) \\ \hline
Upsampling technique & NA & \texttt{torch.repeat\_interleave()}  \\ \hline
\end{tabular}
\caption{Parameters and their values for CIA and SIA modules.}
\label{tab:config-table}
\end{table*}

\begin{figure*}%[!ht]%
\centering
\includegraphics[width=0.9\textwidth]{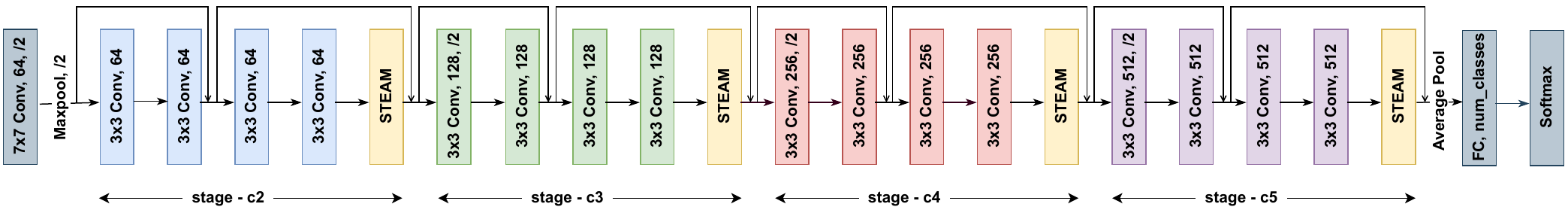}  \hfill
\caption{ResNet-18 integrated with STEAM}
\label{figure:stream-resnet18}
\end{figure*}

 For the placement of STEAM units in ResNets and ShuffleNet-V2, we adopt the same adaptive strategy discussed in the methodology section. To better understand our adaptive strategy, we have illustrated how STEAM is integrated in a standard ResNet-18 (Fig.~\ref{figure:stream-resnet18}). ResNet-18 contains [2, 2, 2, 2] blocks in stages [c2, c3, c4, c5], hence the number of STEAM units for each stage will be \(\lceil 2 / 4 \rceil\ = 1\). Thus, the number of STEAM units per stage for ResNet-18 are [1, 1, 1, 1], implying that only one STEAM unit needs to be added in every stage, hence we position it at the end of each stage. 

\begin{figure}%[!ht]%
\centering
\includegraphics[width=0.45\textwidth]{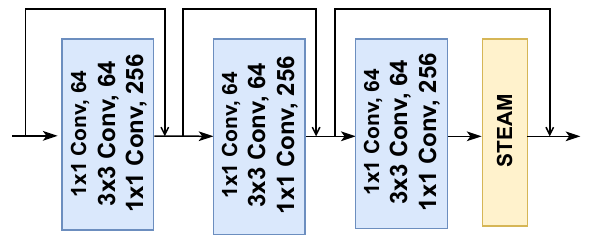}  \hfill
\caption{Stage c2 of ResNet-50 integrated with STEAM.}
\label{stage-c2-resnet50}
\end{figure}

\begin{figure}%[!ht]%
\centering
\includegraphics[width=0.45\textwidth]{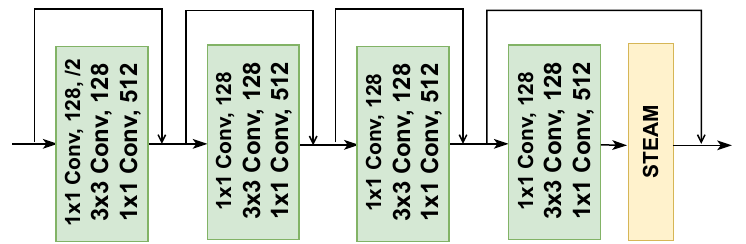}  \hfill
\caption{Stage c3 of ResNet-50 integrated with STEAM.}
\label{stage-c3-resnet50}
\end{figure}

\begin{figure}%[!ht]%
\centering
\includegraphics[width=0.45\textwidth]{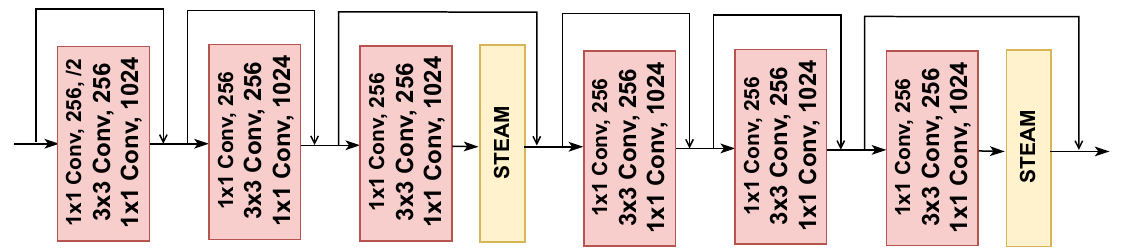}  \hfill
\caption{Stage c4 of ResNet-50 integrated with STEAM.}
\label{stage-c4-resnet50}
\end{figure}

\begin{figure}%[!ht]%
\centering
\includegraphics[width=0.45\textwidth]{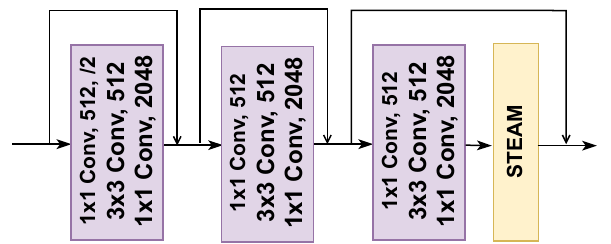}  \hfill
\caption{Stage c5 of ResNet-50 integrated with STEAM.}
\label{stage-c5-resnet50}
\end{figure}

Similarly, ResNet-50 contains [3, 4, 6, 3] blocks in stages [c2, c3, c4, c5], hence the number of STEAM units for each stage will be [\(\lceil 3 / 4 \rceil\), \(\lceil 4 / 4 \rceil\), \(\lceil 6 / 4 \rceil\), \(\lceil 3 / 4 \rceil\)] which equals to [1, 1, 2, 1] respectively. Stages c2, c3, and c5 require only one STEAM unit. Hence we position our module at the end of these stages (Figs. \ref{stage-c2-resnet50}, \ref{stage-c3-resnet50}, \ref{stage-c5-resnet50}).
For stage c4, two STEAM units are needed. To ensure uniform placement, one unit is added at the end of the stage, and the other is placed after the 3rd block (Fig.~\ref{stage-c4-resnet50}). This pattern is consistently applied to ShuffleNet-V2 model as well. Fig.~\ref{alexnet-fig} shows AlexNet integrated with STEAM. 
%We have ommited batchnorm, pooling and activation layers from the Figs. 2-7 to ensure simplicity of images. Readers are encouraged to go through \cite{he2016deep} and \cite{krizhevsky2012imagenet} for exact architecture details of ResNets and AlexNet respectively. 
We have omitted the batch normalization, pooling, and activation layers in Figs. 2-7 to keep the illustrations simple. Readers are encouraged to refer to \cite{he2016deep} and \cite{krizhevsky2012imagenet} for the exact architectural details of ResNets and AlexNet, respectively.

% However, stage c4 requires 2 STEAM units, hence we add one STEAM unit at the end of this stage and the remaining one is positioned after the \(3^{rd}\) block in this stage (Fig.~\ref{stage-c4-resnet50}), which ensures that STEAM units are positioned uniformly within the stage. A similar pattern is followed for ResNet-101 and ShuffleNet-V2 models also.

% \begin{figure}%[!ht]%
% \centering
% \includegraphics[width=0.175\textwidth]{AlexNet.drawio.pdf}  \hfill
% \caption{AlexNet Architecture}
% \label{alexnet-fig}
% \end{figure}

% The value of \(d_{K}\) (used in Eqs. 1 and 4) is not explicitly mentioned but it is quite a common term whose value is equal to \(d / H\) which equals 2 for both CIA and SIA.
% Regarding placement of STEAM units in ResNets and ShuffleNet-V2, we follow the same adaptive strategy as discussed in our methodology section.
% \textbf{NOTE: } we follow the same configuration for STEAM across all ablations performed in main paper and in Appendix. Only those parameters are changed for which the ablation is being performed. 

\section{Ablations}

\begin{table*}[h]
% \small
  \centering
  \scriptsize
\begin{tabular}{c|c|c|c|c|c|c|c|c|c}
\hline
\multicolumn{10}{c}{\textbf{ImageNet-100 Classes}} \\ \hline
n02869837 & n01749939 & n02488291 & n02107142 & n13037406 & n02091831 & n04517823 & n04589890 & n03062245 & n01773797 \\ \hline
n01735189 & n07831146 & n07753275 & n03085013 & n04485082 & n02105505 & n01983481 & n02788148 & n03530642 & n04435653 \\ \hline
n02086910 & n02859443 & n13040303 & n03594734 & n02085620 & n02099849 & n01558993 & n04493381 & n02109047 & n04111531 \\ \hline
n02877765 & n04429376 & n02009229 & n01978455 & n02106550 & n01820546 & n01692333 & n07714571 & n02974003 & n02114855 \\ \hline
n03785016 & n03764736 & n03775546 & n02087046 & n07836838 & n04099969 & n04592741 & n03891251 & n02701002 & n03379051 \\ \hline
n02259212 & n07715103 & n03947888 & n04026417 & n02326432 & n03637318 & n01980166 & n02113799 & n02086240 & n03903868 \\ \hline
n02483362 & n04127249 & n02089973 & n03017168 & n02093428 & n02804414 & n02396427 & n04418357 & n02172182 & n01729322 \\ \hline
n02113978 & n03787032 & n02089867 & n02119022 & n03777754 & n04238763 & n02231487 & n03032252 & n02138441 & n02104029 \\ \hline
n03837869 & n03494278 & n04136333 & n03794056 & n03492542 & n02018207 & n04067472 & n03930630 & n03584829 & n02123045 \\ \hline
n04229816 & n02100583 & n03642806 & n04336792 & n03259280 & n02116738 & n02108089 & n03424325 & n01855672 & n02090622 \\ \hline
\end{tabular}
\caption{100 classes randomly sampled from ImageNet to create ImageNet-100 dataset.}
  \label{tab:imagenet100}
\end{table*}
In this section, we conduct extensive ablations to identify the optimal mechanisms and hyperparameters for designing our proposed method, STEAM: \textit{Squeeze and Transform Enhanced Attention Module}.
For these ablations, we use ResNet-18 as a backbone integrated with STEAM and evaluate its effectiveness on image classification using a subset of the ImageNet dataset, consisting of 100 randomly selected classes, forming the ImageNet-100 subset (as shown in Table \ref{tab:imagenet100}), with approximately 128K training images and 5K validation images. The selected 100 classes are the same as those used by \cite{wang2020understanding}. We report Top-1 and Top-5 accuracy for all ablations in Tables 3-11 where bolded values denote the best performance.

We study the following ablations:
\begin{itemize}
    \item Effect of different pooling techniques used in CIA module (Table \ref{pool-cia}).
    \item Effect of different pooling techniques used in SIA module (Table \ref{pool-sia}).
    \item Effect of random edge drop in SIA (Table \ref{edge-drop}).
    \item Effect of different activation functions used between CIA and SIA (Table \ref{activation-func}).
    \item Effect of \(d\) and \(H\) on STEAM (Table \ref{outdim-heads}).
    \item Comparison of STEAM with it's SOTA counterparts on ImageNet-100 (Table \ref{sota}).
    \item Effect of Output Guided Pooling on CBAM (Table \ref{cbam-ogp}).
    \item Comparision of na\"ive channel attention with our proposed CIA module (Table \ref{knn}).
    \item Comparision of STEAM with it's SOTA counterparts on ImageNet-100 using a lightweight model, AlexNet (Table \ref{alexnet}).
\end{itemize}

\subsection{Implementation Details}  
We follow the same adaptive strategy for integrating STEAM into ResNet-18, adding SE, CBAM, ECA, and GCT modules before each residual connection.
All the models were trained on a single NVIDIA RTX 4090 GPU using a mini-batch size of 256. We applied the same training procedure to all models, including random cropping to \(224 \times 224\) and random horizontal flipping. We use the \textit{Stochastic Gradient Descent} (SGD) optimizer with a momentum of 0.9 and a weight decay of 1e-4. All models were trained for 50 epochs with an initial learning rate of 0.1, which was decremented by a factor of 10 at \(20^{th}\), \(40^{th}\) epochs. While evaluating the models on the validation set, we begin by resizing the input image to \(256 \times 256\) followed by a centre crop of \(224 \times 224\) and random horizontal flipping. \textbf{NOTE: } We follow the same training configuration for all the ablation experiments unless otherwise specified. 

 % In this section we study different basic blocks of STEAM unit comprehensive and find empirical results to support our claims and further investigate the effects of various different components in it. ALl the experiments were performed using ResNet18 backbone and Imagenet100 dataset at 50 epochs.

\textbf{Pooling techniques in CIA}
First, we examine the impact of various pooling mechanisms in the CIA, specifically global average pooling, global max pooling, and a combination of both through concatenation. As shown in Table \ref{pool-cia}, global average pooling results in slightly better Top-1 accuracy compared to the concatenation of global average and max pooling. However, global max pooling produced suboptimal results.
\begin{table}[h]
  \centering
  \begin{tabular}{c|c|c}
    \hline
    Pooling Mechanism & Top-1 & Top-5 \\ \hline
    Global Average Pool (GAP) & \textbf{81.14} & 94.79  \\ \hline
    Global Max Pool (GMP)  & 80.57 & \textbf{94.82}  \\ \hline
    Concatenation (GAP \& GMP) & 80.85 & 94.73 \\ \hline
  \end{tabular}
  \caption{Effect of different pooling techniques used in CIA.}
  \label{pool-cia}
\end{table}

 \textbf{Pooling techniques in SIA}
Next, we assess the effects of various pooling mechanisms in SIA, including global max pooling, global average pooling, and their combination via concatenation. Consistent with the CIA results, global average pooling yields slightly better Top-1 accuracy compared to the other approaches. The results of this ablation are presented in Table \ref{pool-sia}.
\begin{table}[h]
  \centering
  \begin{tabular}{c|c|c}
    \hline
    Pooling Mechanism & Top-1 & Top-5 \\ \hline
    Global Average Pool (GAP) & \textbf{81.14} & 94.79  \\ \hline
    Global Max Pool (GMP) &  80.46 & 94.70  \\ \hline
    Concatenation (GAP \& GMP) & 80.95 & \textbf{94.85}  \\ \hline
  \end{tabular}
  \caption{Effect of different pooling techniques used in SIA.}
  \label{pool-sia}
\end{table}

\textbf{Effect of random edge drop in SIA}
As discussed in the methodology, we hypothesized that several nodes having a uniform degree of 4 within the central submatrix of our refined spatial map could lead to over-smoothing, resulting in identical node representations and degrading the model’s performance. To investigate this, we performed two ablations: one without random edge drop and another with random edge drop applied specifically to nodes in the central submatrix within SIA. Table \ref{edge-drop} demonstrates that applying random edge drop to these nodes does, in fact, improve the model's performance. 
\begin{table}[h]
  \centering
  \begin{tabular}{c|c|c}
    \hline
    Edge Drop in SIA & Top-1 & Top-5 \\ \hline
    No   & 80.58 & 94.36  \\ \hline
    Yes   & \textbf{81.14} & \textbf{94.79}   \\ \hline
  \end{tabular}
  \caption{Effect of random edge drop in SIA.}
  \label{edge-drop}
\end{table}

\textbf{Different activation functions}
Next, we examine the impact of different non-linearities (activations) used between our CIA and SIA modules. We hypothesize that introducing non-linearity between these two modules can enhance the model’s ability to capture more complex relationships. Empirical results, as shown in Table \ref{activation-func}, clearly indicate that incorporating any non-linearity is better than none, leading to improved Top-1 accuracy. Among ReLU, Tanh, and Sigmoid, we find that Tanh yields the best results.
\begin{table}[h]
  \centering
  \begin{tabular}{c|c|c}
    \hline
   Non-linearity used & Top-1 & Top-5 \\ \hline
    Nothing   & 80.31	& 94.48   \\ \hline
    ReLu   & 80.52 & 94.71  \\ \hline
    Tanh   & \textbf{81.14 }& \textbf{94.79} \\ \hline
    Sigmoid   &80.64 & 94.69  \\ \hline
  \end{tabular}
  \caption{Effect of different non-linearity used between CIA and SIA.}
  \label{activation-func}
\end{table}

\textbf{Effect of \(d\) and \(H\) on STEAM}
Next, we discuss the effect of \(d\), the hidden dimension and \(H\), the number of heads used in our multi-head graph attention. We experiment with the following pairs of (\(d\), \(H\))s: (4, 1), (4, 2), (8, 1), (8, 4), (8, 8), (16, 1), (16, 4), (16, 8). Table \ref{outdim-heads} shows that using (\(d=8\), \(H=4\)) yields slightly better Top-1 accuracy than (\(d=8\), \(H=8\)) but the latter achieves better Top-5 accuracy. 

\begin{table}[h]
  \centering
  \begin{tabular}{c|c|c|c}
    \hline
    \(d\) & \(H\) & Top-1 & Top-5 \\ \hline
    4 & 1  & 80.45   & 94.61   \\ \hline
    4 & 2  & 80.57   & 94.64  \\ \hline
    8 & 1  & 80.75  & 94.69  \\ \hline
    8 & 4 & \textbf{81.14}   & 94.79   \\ \hline
    8 & 8 & 81.02  & 94.88 \\ \hline
    16 & 1  & 80.61   & 94.70  \\ \hline
    16 & 4  & 80.98  & \textbf{94.90}  \\ \hline
    16 & 8   & 80.77  & 94.67 \\ \hline

  \end{tabular}
  \caption{Effect of \(d\) and \(H\) on STEAM.}
  \label{outdim-heads}
\end{table}

\textbf{Comparison with SOTA modules on ImageNet-100}
After identifying the optimal parameters for our module, we proceed to compare STEAM with its SOTA counterparts—SE, CBAM, ECA, and GCT. As shown in Table \ref{sota}, STEAM outperforms all these leading modules on our ImageNet-100 dataset, demonstrating the effectiveness of our approach even on smaller datasets. Specifically, STEAM achieves 1.34\%, 0.82\%, 0.49\%, 0.66\%, and 0.29\% higher Top-1 accuracy than the standard ResNet-18, SE, CBAM, ECA, and GCT methods, respectively.
\begin{table}[h]
  \centering
  \begin{tabular}{c|c|c}
    \hline
    Method & Top-1 & Top-5 \\ \hline
    ResNet-18   & 79.80 & 94.27 \\ \hline
     +SE   &  80.32  & 94.52  \\ \hline 
     +CBAM   & 80.65  & 94.67 \\ \hline
     +ECA   & 80.48   & 94.58 \\ \hline
     +GCT   & 80.85   & 94.73 \\ \hline
     +STEAM  &  \textbf{81.14} & \textbf{ 94.79} \\ \hline
  \end{tabular}
  \caption{Image classification results on ImageNet-100 using ResNet-18 as backbone.}
  \label{sota}
\end{table}

\begin{figure}%[!ht]%
\centering
\includegraphics[width=0.14\textwidth]{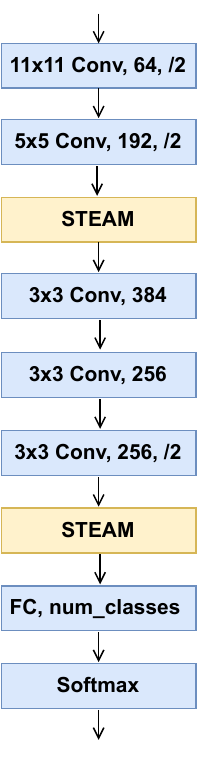}  \hfill
\caption{AlexNet integrated with STEAM}
\label{alexnet-fig}
\end{figure}

\textbf{Effect of Output Guided Pooling}
Next, we analyze the effectiveness of our proposed \textit{Output Guided Pooling (OGP)}. OGP transforms the intermediate feature map \(\mathbb{R}^{C \times H \times W}\) into a fixed-size spatial map (\(\in \mathbb{R}^{m \times m}\)) using global average pooling (GAP) along the channel dimension, where \(m\) corresponds to the spatial dimension of the deepest layer. For both ResNet and ShuffleNet-V2 models, \(m = 7\). Applying spatial interaction attention (SIA) directly to the intermediate feature map \(\mathbb{R}^{C \times H \times W}\) would result in \(H*W\) nodes, which can be computationally expensive for early layers and lead to unnecessary overhead (as discussed in the methodology). To evaluate the effectiveness of OGP, we integrate it into the spatial attention module of CBAM and train the model from scratch. Similar to STEAM, the spatial attention module of CBAM now generates spatial scores \(\in \mathbb{R}^{m \times m}\), which are then upsampled using \texttt{torch.repeat\_interleave()} to produce the final spatial scores \(\in \mathbb{R}^{C \times H \times W}\). Table \ref{cbam-ogp} shows that incorporating OGP into the spatial attention module of CBAM results in a 0.22\% improvement in Top-1 accuracy compared to the standard CBAM.

% Next we analyze the effectiveness of our proposed \textit{Output Guided Pooling} (OGP). OGP transforms the given intermediate feature map \(\mathbb{R}^{C \times H \times W}\) into a fixed-size spatial map (\(\in \mathbb{R}^{(m \times m)}\)) using GAP along the channel dimension, where \(m\) corresponds to the dimension of the deepest spatial map. For both ResNets and ShuffleNet-V2, \(m = 7\). Applying SIA directly onto the intermediate feature map \(\mathbb{R}^{C \times H \times W}\) would translate to \(H*W\) nodes which can be very high for early layers and lead to unnecesary computations (as explained in methodology). Hence, to test the effectiveness of OGP, we apply it to the spatial attention module of CBAM and train the model from scratch. Similar to STEAM, the spatial attention module of CBAM now generates spatial scores \(\in \mathbb{R}^{m \times m}\) which are upsampled using \texttt{torch.repeat\_interleave()} to produce the final spatial scores \(\in \mathbb{R}^{C \times H \times W}\). Table \ref{cbam-ogp} shows that incorporating OGP in spatial attention module of CBAM achieves 0.22\% higher Top-1 accuracy when compared to standard CBAM.
\begin{table}[h]
  \centering
  \begin{tabular}{c|c|c}
    \hline
    Method & Top-1 & Top-5 \\ \hline
    CBAM & 80.65 & 94.67 \\ \hline
    CBAM with OGP &\textbf{ 80.87} & \textbf{94.72} \\ \hline
  \end{tabular}
  \caption{Effect of OGP on CBAM.}
  \label{cbam-ogp}
\end{table}

\textbf{Comparision with our na\"ive channel attention module}
As described in the methodology, due to the challenges of using dense graphs, we experimented with \(k\)-NN graphs with \(k=16, 32, 64 \). For these experiments, we used ResNet-18 as the backbone and integrated only the \(k\)-NN graph (we refer to this configuration as the \(k\)-CAM : \(k\)-Channel Attention Module) and our proposed channel attention module, excluding the SIA module. As shown in Table \ref{knn}, the \(k\)-NN channel attention module (across different values of \(k\)) delivers suboptimal results compared to our proposed CIA module.
\begin{table}[h]
  \centering
  \begin{tabular}{c|c|c}
    \hline
    Module & Top-1 & Top-5 \\ \hline
    16-CAM (\(k\)=16)  & 80.11  & 94.42 \\ \hline
    32-CAM (\(k\)=32)   &  80.18  & 94.45  \\ \hline
    64-CAM (\(k\)=64)   &  79.96 & 94.31   \\ \hline
    CIA   & \textbf{80.56 }& \textbf{94.80 }  \\ \hline
  \end{tabular}
  \caption{Comparision of na\"ive channel attention and our proposed CIA module.}
  \label{knn}
\end{table}

\textbf{Comparision with a lightweight model}
Next, we evaluate the effectiveness of our module with a lightweight model, AlexNet, on the ImageNet-100 dataset. We use a standard AlexNet but replace its two fully connected layers with a single fully connected layer. As shown in Fig.~\ref{alexnet-fig}, AlexNet is integrated with STEAM, with two STEAM units added after the second and fifth convolutional layers. We replace STEAM with SE, CBAM, ECA, and GCT modules when training these respective models. For the standard AlexNet, these modules are simply omitted during training. Table \ref{alexnet} shows that STEAM outperforms all its SOTA counterparts even on a lightweight model like AlexNet.
\begin{table}[h]
  \centering
  \begin{tabular}{c|c|c}
    \hline
    Method & Top-1 & Top-5 \\ \hline
    ResNet-18   & 65.74 & 87.02 \\ \hline
     +SE   &  66.19  & 87.61 \\ \hline 
     +CBAM   & 66.63  & \textbf{88.24} \\ \hline
     +ECA   & 66.25 & 87.60 \\ \hline
     +GCT   & 67.02 & 88.14 \\ \hline
     +STEAM  &  \textbf{67.26} & 87.95 \\ \hline
  \end{tabular}
  \caption{Image classification results on ImageNet-100 using AlexNet as backbone.}
  \label{alexnet}
\end{table}
\end{document}